\documentclass[5p, times]{elsarticle}

\usepackage{ecrc}
\usepackage[inline]{enumitem}
\usepackage{comment}
\usepackage{graphicx} 
\usepackage{amsmath} 
\usepackage{amsfonts}
\usepackage{mathtools}
\usepackage{microtype}
\usepackage{subfig}
\usepackage{tikz}
\usepackage{adjustbox}
\usepackage{booktabs}
\usepackage{multirow}
\usepackage{url}
\usepackage{svg}
\usepackage{optidef}
\usepackage{xcolor,colortbl}

\usepackage{algorithm}

\usepackage{algpseudocode}
\usepackage{hyperref}
\hypersetup{
 colorlinks=true,
 linkcolor=blue,
 filecolor=magenta, 
 urlcolor=cyan,
 citecolor= gray,
 pdftitle={Overleaf Example},
 pdfpagemode=FullScreen,
 pdfauthor=whatever
 }

\usepackage[separate-uncertainty=true]{siunitx}
\usepackage{dirtytalk}

\usepackage{threeparttable} 

\usepackage{colortbl}
\usepackage{multirow}
\usepackage{makecell}

\usepackage{romannum}

\usepackage{flushend}
\usepackage{tabularx}

\usepackage{soul} 

\newcommand{\ie}{\textit{i.e.}}
\newcommand{\eg}{\textit{e.g.}}

\usepackage[normalem]{ulem} 

\volume{00}

\firstpage{1}

\journalname{Accepted for publication inRobotics and Computer-Integrated Manufacturing (2025)}

\runauth{S. Sandrini et al.}

\usepackage{amssymb}

\definecolor{Gray}{gray}{0.95}
\definecolor{LightGray}{gray}{0.95}

\definecolor{LightCyan}{rgb}{0.88,1,1}

\newcolumntype{G}{>{\columncolor{Gray}}c}
\newcolumntype{L}{>{\columncolor{white}}c}

\begin{document}

\begin{frontmatter}

 \title{ Learning and planning for optimal synergistic human-robot coordination in manufacturing contexts}

 \author[STIIMA,POLITO]{Samuele Sandrini\corref{cor1}}\ead{samuele.sandrini@stiima.cnr.it}
 \author[POLIMI]{Marco Faroni}
 \author[STIIMA]{Nicola Pedrocchi}

 \cortext[cor1]{Corresponding Author.}

 \address[STIIMA]{Institute of Intelligent Industrial Technologies and Systems for Advanced Manufacturing, National Research Council of Italy, 20133, Milan, Italy}
 \address[POLIMI]{Dipartimento di Elettronica, Informazione e Bioingegneria, Politecnico di Milano, Milan, Italy}
 \address[POLITO]{Department of Control and Computer Engineering, Politecnico di Torino, Torino, Italy}

 \begin{abstract}
 {Collaborative robotics cells} leverage heterogeneous agents to provide agile production solutions. 
 {Effective coordination is essential to prevent inefficiencies and risks for human operators working alongside robots.}
 This paper proposes a human-aware task allocation and scheduling model based on Mixed Integer Nonlinear Programming to optimize efficiency and safety  {starting from task planning stages}.
 The approach exploits synergies that encode the coupling effects between pairs of tasks executed in parallel by the agents{, arising from the safety constraints imposed on robot agents.}
 These terms are learned from previous executions using a Bayesian estimation; the inference of the posterior probability distribution of the synergy coefficients is performed using the Markov Chain Monte Carlo method. The synergy enhances task planning by adapting the nominal duration of the plan according to the effect of the operator's presence.
 {Simulations and experimental results demonstrate that the proposed method produces improved human-aware task plans, reducing unuseful interference between agents, increasing human-robot distance, and achieving up to an 18\% reduction in process execution time.}

 \end{abstract}

\end{frontmatter}

\section{Introduction}\label{sec: intro}

\begin{figure*}[t]
 \centering
 \includegraphics[trim={0cm 0cm 0cm 0cm},clip,width=0.85\textwidth]{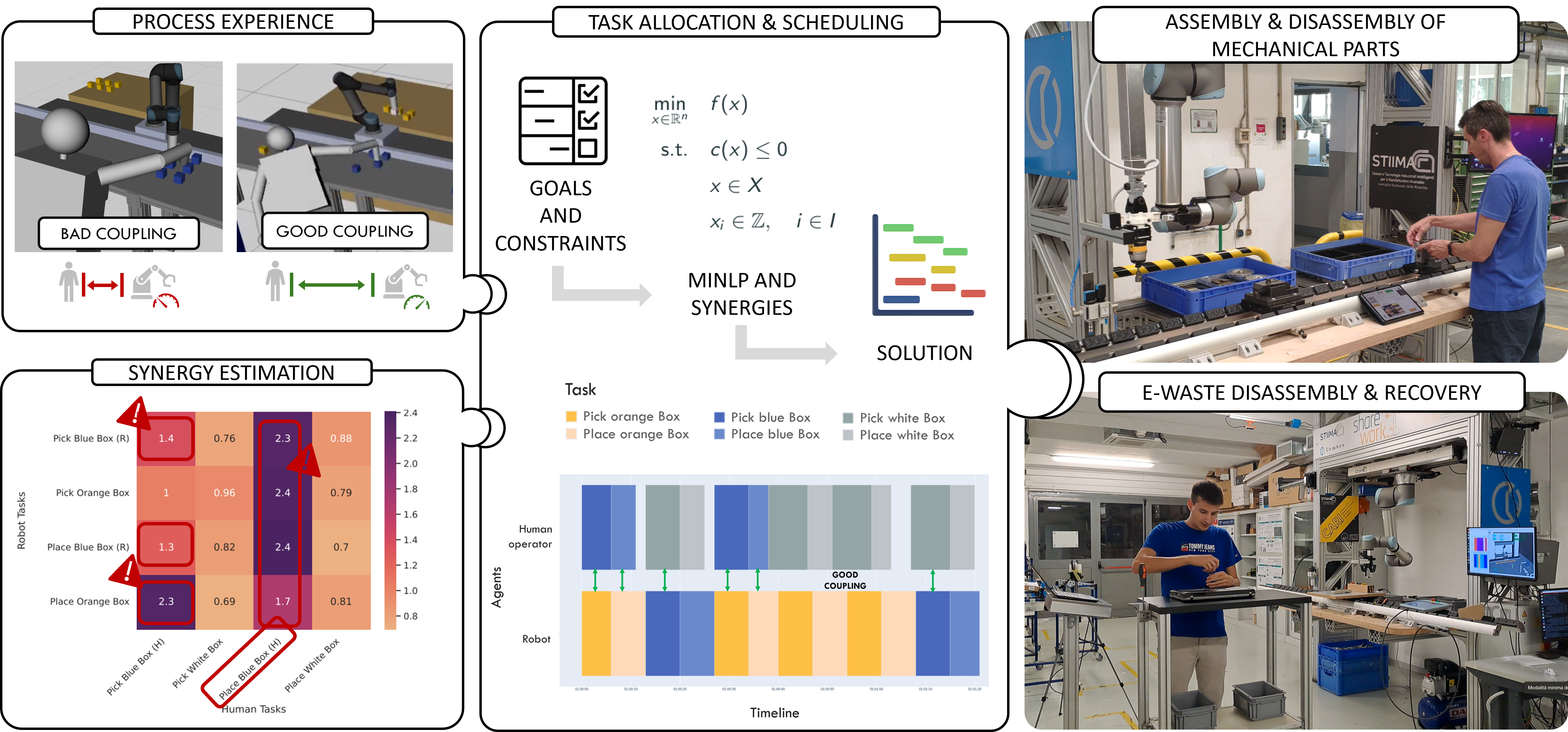}
 \caption{Visual description of the proposed method. The boxes on the left show how task statistics and task synergies of the various agents are estimated using process experience, highlighting good or bad couplings between them. In the center, it is shown how these synergies are integrated to improve the MINLP algorithm for task allocation and scheduling. The boxes on the right show two industrial application scenarios that can benefit from the proposed method. Supplementary material can be found: \url{https://jrl-cari-cnr-unibs.github.io/synergistic_hrtp/}.
 }\label{fig:visual_description}
\end{figure*}

Production stations often involve the coexistence of human operators and robots to ensure flexibility, reconfigurability, and suitability for high-mix, low-volume production demands.
Safe\-ty has a direct impact in terms of efficiency in collaborative environments~\cite{hrc_safety_kpi_risk_assessment}, \eg, the \emph{Speed and Separation Monitoring} (SSM) proposed by ISO/TS 15066~\cite{ISOTS15066} requires that the robot speed be modulated based on the agents' distance to keep a protective separation.
Researchers have tackled the ``safety-vs-efficiency'' trade-off by mainly acting on motion planning and control levels (e.g., impedance control~\cite{franceschi2022adaptive, song2019tutorial}, safety-aware path planning~\cite{hamp_faroni, s_star_hadddadin}, and fast motion replanning~\cite{tonola2023anytime}). At the same time, only a few works focused on the task planning layer (e.g.~\cite{faroni_optimal_tamp}).
When the task planner is unaware of safety aspects, it produces plans that may lead to interference between the agents and increase the frequency of safety stops that must be activated to guarantee the operators' safety.
In such a context, our paper investigates the coordination of agents to achieve efficient production, considering safety-related effects and agent interference. The approach pursues a {synergistic} human-robot collaboration for Industry 5.0{~\cite{demir2019industry}}, with particular relevance to assembly and disassembly applications, object sorting, and bin picking for collaborative robotic cells, as shown in Figure~\ref{fig:visual_description}.

\subsection{Related Works}

Task Allocation and Scheduling in Human-Robot Collaboration (HRC) have been tackled using heterogeneous approaches.
Several works focused only on the task allocation problem~\cite{task_allocation_review_in_hrc}.
For example,~\cite{johannsmeier2016hierarchical} uses AND/OR graph plan representation~\cite{de1990and} and solves the optimal allocation problem by graph search with A*.
Building upon this work,~\cite{lamon2019capability} emphasized the agent's capability-based allocation problem by introducing indices about task complexity, agent dexterity, and effort.
In~\cite{merlo2022dynamic} and~\cite{MERLO2023111}, the problem of assigning tasks to agents online is addressed using the AND/OR graph representation and the AO* graph search, taking into account the physical load of the agent that is quantified by motion capture and ergonomic risk assessment. The ergonomics factor in the allocation problem plays a crucial role in~\cite{ busch2018planning, makrini2019task}. In~\cite{fusaro2021integrated}, the online task allocation problem is integrated with a custom \emph{control node} into a \emph{Behavior Tree}~\cite{iovino2022survey}; the allocation is solved through Mixed Integer Linear Programming (MILP) on a subset of tasks at runtime only when it is time to perform that subset of actions.
\cite{monguzzi2022mixed} investigates both allocation and scheduling. The allocation is solved offline based on agent capabilities and is decoupled from the scheduling. A variant of the Hungarian algorithm~\cite{hungarian_algorithm} solves the allocation, while a MILP approach solves the scheduling.

Some works incorporate the degradation of human performance during execution~\cite{degradation_performance, pupa2021dynamic, zhang2020real}.
The work in~\cite{pupa2021human} introduces the concept of awareness of the actual duration of the tasks performed by the operator using a two-level architecture: (i) a task allocation and scheduling layer based on Multi-Objective MILP and (ii) a dynamic scheduling layer that reschedules when triggered by the agents monitoring.
The same architecture was used in~\cite{pupa2022resilient} to adapt the schedule locally in case of failure.

In~\cite{faroni2020layered} and~\cite{faroni_optimal_tamp} the human-robot coordination is pursued through a \emph{Timeline-based} task planner included in the \emph{Task and Motion Planning (TAMP)} framework~\cite{garrett2021integrated}.\@
In~\cite{marc_toussaint_hrc_dlgp}, the authors combine TAMP with a human motion prediction module based on a goal-conditioned recurrent neural network to reduce human-robot interference.
\emph{Human-Aware Task Planning} was studied by Alami et al.\ in~\cite{alami2006toward} using the \emph{Hierarchical Task Network} paradigm~\cite{georgievski2015htn}.
This evolved into the \emph{HATP (Hierarchical Agent-based Task Planner)} framework~\cite{de2015hatp, lallement2014hatp, lallement2018hatp} and its variants~\cite{buisan2021human},
which consider the human agent as a rational agent with its decision model.
\emph{Theory of Mind (ToM)} was recently explicitly introduced at the task planning level in~\cite{false_beliefs_TOM_TP}.

A branch of the literature focused on allocation and scheduling problems for industrial applications.
For instance,~\cite{ham2021human} handles the coordination of heterogeneous agents using a MILP formalization and constrained scheduling for aircraft assembly;~\cite{umbrico2022enhanced} and~\cite{umbrico2022design} apply timeline-based planning in the TAMP framework to a semi-automated assembly/disassembly industrial case study; 
~\cite{liu2022task} proposed a task allocation technique for HRC using a neural network for electric vehicle battery disassembly;~\cite{bogner2018optimised} takes up the MILP formulation to handle allocation and scheduling for the assembly of printed circuit boards using meta-heuristic techniques; and~\cite{liau2020task} solves the allocation problem using genetic algorithms for HRC in mold assembly processes.

Some works consider safety and efficiency jointly at task planning level.~\cite{lippi2021mixed} formalizes human-multi-robot allocation and scheduling problems jointly based on MILP; the cost function is optimized by considering the makespan, an index of task execution quality, and the agents' workload.
They consider safety issues by introducing \emph{spatial constraints}, avoiding the parallel execution of tasks in the same work area, and providing online monitoring and rescheduling. The agents' occupancy maps are known beforehand.
In~\cite{lippi2022task}, they introduce the cost of task switches during online rescheduling.
~\cite{pupa2021safety} reformulates the dynamic scheduler in~\cite{pupa2021dynamic, pupa2021human} to online adapt the schedule locally in case of safety halts. 
However, this implements an online countermeasure to risky or blocking situations without the proactivity needed to avoid such situations. 
~\cite{faccio2023task} considers safety by parameterizing the optimization problem with the nominal SSM parameters. It assumes that each task pair is associated with a pair of 2D points and corresponds to a safety speed reduction, and the human movement is static at that point and deterministic given a task. 

\begin{table*}[t]
 \caption{List of main variables involved in model formulation for the human-aware task allocation and scheduling.}\label{table: list_of_notations}
 \centering
 \newcolumntype{R}{>{\raggedleft\arraybackslash}p{0.16\textwidth}}
 \begin{tabularx}{\textwidth}{RX}
 \toprule
 \textbf{Symbol} & \textbf{Description} \\
 \midrule
  $\mathcal{T}$, $m=|\mathcal{T}|$, $\tau_i\in\mathcal{T} $ & The task set, the total number of tasks, the $i$-th task \\

 $\mathcal{A^R}$, $H$, $\mathcal{A} = \mathcal{A}^R \cup H$  & The robot agent set, the human agent, the set of agents \\

 $t^s_i \in \mathbb{R}^+$ and $t^e_i \in \mathbb{R}^+$  & Start-Time and End-Time of task $\tau_i$ \\
 $a^j_i \in \left\{0,1\right\}$ & Binary variable for the assignment of task $\tau_i$ to agent $j$ \\
 $C_{i,j} \in \left\{0,1\right\}$ & Binary input variable that specifies whether agent $j$ can perform task $\tau_i$ \\
 $P_{i,k} \in \left\{0,1\right\}$ & Binary input variable that specifies whether task $\tau_i$ must precede task $\tau_k$ \\
 $\hat{d}^j_i \in \mathbb{R}^+$ & Expected duration of task $\tau_i$ executed by agent $j$ \\
 $\mathrm{OV}_{i,k} \in \mathbb{R}_0^+ $ & Overlapping time between tasks $\tau_i$ and $\tau_k$ \\
 $s^j_{i,k} \in \mathbb{R}^+$ & Synergy term between task $\tau_i$ (executed by agent $j$) and task $\tau_k$ (executed by the human agent) \\
 $\mathcal{D}$ & Dataset containing task observations\\

 \bottomrule
 \end{tabularx}
\end{table*}

\subsection{Contribution}\label{sec: contribution}

Although most of the literature considers safety as an online countermeasure, {this paper takes a synergistic approach at the task planning level by introducing synergy coefficients into the planning model. This provides a safe and efficient solution to agents' coordination in collaborative robotics.}
The key contributions of this paper can be summarized as follows:
\begin{description}[leftmargin=6pt]
\item[Learning of human-robot synergies.] We estimate the coupling effects on execution time between pairs of human-robot tasks, leveraging knowledge gained from previous process executions. Building on the definition of synergy that we presented in~\cite{sandrini2022learning}, we adopt a Bayesian estimation approach to learn coefficients. 
This approach iteratively updates the estimated synergy distribution according to new process observations.
\item[Human-aware task planning.] Mixed Integer Nonlinear Programming (MINLP) is used {to define human-aware task allocation and scheduling models}. It integrates the learned synergies into the planning model to consider the coupling effect (detrimental or beneficial) caused by the parallel execution of tasks between robots and operators. We proposed two models: the first is called Synergistic Task Planning (STP), in which the completion time of robots' tasks is interdependent through synergy values with tasks executed in parallel by the human agent, and the second, Relaxed Synergistic Task Planning (R-STP), which reduces the computational complexity of STP by considering a simplified synergy term in the cost function.
\end{description}

Unlike state-of-the-art approaches based on Mixed Integer Linear Programming, {in the proposed models, the execution times of tasks performed by robotic agents are not fixed to their nominal values independently of human operator actions executed in parallel. Instead, they are influenced by these actions and adjusted according to synergy values. Moreover, the proposed approach is not parameterized over low-level implementations of safety specifications, geometric task layouts, or manually defined human-robot interactions.}
This makes our approach flexible for reconfiguring the process and robot programming. 
{Such reconfiguration requires learning the synergy values and tasks' duration from new simulated or actual process data.} 
%
Compared to~\cite{faccio2023task} and~\cite{lippi2021mixed}, our models do not rely on the nominal parameters of the SSM, nor \emph{a-priori} definition of static points associated with tasks, nor on manually tuned constraints of non-parallelism between tasks heuristically considered close.

The proposed methods were evaluated and compared with multiple baselines in a simulated HRC scenario and a case study of electronic waste disassembly in the real world. 
Simulations and real-world experiments demonstrate that our approach reduces the cycle time (up to \SI{18}{\percent}) and leads to safer executions with a larger average human-robot distance. 
A {preliminary} human factor analysis also revealed a preference for the proposed methods in terms of robot proximity, interruptions, and overall satisfaction.  
Video of the experiments is available online: \href{https://youtu.be/GGVBM0dyQYo}{https://youtu.be/GGVBM0dyQYo}.
We show how to linearize MINLP models in~\ref{appendix: a}.

\section{Problem Formulation}\label{sec: methodology}

The allocation and scheduling of all the tasks $\tau_i \in \mathcal{T}$ to the available agents $j \in \mathcal{A}$ in order to obtain the optimal plan $\pi^*$ that minimizes the makespan of the process can be formalized as an MINLP problem as follows:
\begin{argmini}
 {t^s_i, t^e_i, a^j_i}{\mathcal{M}}
 {\label{milp_optimization_problem}}{\pi^*=}
 \addConstraint{\mathcal{C}_{\mathrm{goal}}}
 \addConstraint{\mathcal{C}_{\mathrm{precedence}}}
 \addConstraint{\mathcal{C}_{\mathrm{capability}}}
 \addConstraint{\mathcal{C}_{\mathrm{overlapping}}}
 \addConstraint{\mathcal{C}_{\mathrm{performance}}}
\end{argmini}
where $\mathcal{M}$ is the makespan:
\begin{equation}\label{makespan_cost_function}
 \mathcal{M} = \max_{i=1,\dots,m}{\left(\,t^e_i\,\right)},
\end{equation}
{where $t^s_i$ and $t^e_i$ are the start and end times of the tasks, respectively, and $a^j_i$ represents the allocation variable of tasks to agents.} The optimization constraints are discussed in the following paragraphs. 

Please note that all variables involved are listed in Table \ref{table: list_of_notations}.

\subsection{Process Constraints}

$\mathcal{C}_{\text{goal}}$ imposes that all the tasks involved in the process shall be executed once:
\begin{equation}\label{c_goal}
 \mathcal{C}_{\mathrm{goal}}:= \quad \sum_{j \in \mathcal{A}}{a^j_i=1} \qquad \forall i \in \left\{1,..., m \right\}.
\end{equation}

\subsection{Precedence Constraints}

The constraint $\mathcal{C}_{\mathrm{precedence}}$ enforces precedence between tasks by imposing that the start time of $\tau_i$ is greater than the end time of {$\tau_k$}. Specifically, denote $P_{i,k}$ as a boolean variable that models the precedence between task $\tau_i$ and $\tau_k$: 

\begin{equation}
 P_{i,k}=\left\{
 \begin{aligned}
 & 1 &\parbox{20em}{if $\tau_i$ must be executed before $\tau_k$,} \\
 & 0 &\parbox{20em}{otherwise (there is no order between $\tau_i$ and $\tau_k$).}
 \end{aligned}
 \right.
\end{equation}
Then, the $\mathcal{C}_{\mathrm{precedence}}$ can be expressed as:
\begin{equation}
 \mathcal{C}_{\mathrm{precedence}}:= 
\left\{ 
	\begin{array}{l}
	\ldots\\
	t^s_i\geq t^e_k P_{i,k}\\
	\ldots\\
	\end{array}
\right. 
\quad \forall i,k \;\;\vee\;\; k \neq i.
\end{equation}

\subsection{Capability Constraints}

$\mathcal{C}_{\mathrm{capability}}$ prevents a task is assigned to an agent that cannot perform it. Specifically, denote $C_{i,j}$ as a boolean variable that gathers the possible assignment of task $\tau_i$ to the $j$-th agent:
\begin{equation}
 C_{i,j}=\left\{
 \begin{aligned}
 & 1 & \parbox{12em}{if agent $j$ can execute $\tau_i$,} \\
 & 0 & \parbox{12em}{otherwise.}
 \end{aligned}
 \right.
\end{equation}
Thus, the capability constraint can be defined as: 
\begin{equation}\label{c_capability}
 \mathcal{C}_{\mathrm{capability}}:= 
 \left\{ 
	\begin{array}{l}
	\ldots\\
	a^j_i \leq C_{i,j}\\
	\ldots\\
\end{array}
\right. 
\quad \quad \forall i,j.
\end{equation}

\subsection{Non-overlapping Constraints}

$\mathcal{C}_{\mathrm{overlapping}}$ avoids overlapping of tasks assigned to the same agent.
Specifically, we impose: 
\begin{equation}\label{s_i_and_s_k}
	\begin{split}
	\mathcal{C}_{\mathrm{overlapping}}:= 
\left\{ 
\begin{array}{l}
\ldots\\
t^s_i > t^s_k - M (1-\delta_{i,k}) - M (2-a^j_i-a^j_k) \\
t^s_i \leq t^s_k + M \delta_{i,k} + M (2-a^j_i-a^j_k) \\
t^e_i > t^e_k - M (1-\delta_{i,k}) - M (2-a^j_i-a^j_k)\\
t^e_i \leq t^s_k + M \delta_{i,k} + M (2-a^j_i-a^j_k) \\
 \ldots
\end{array}
\right.\\
\forall i,k \quad \vee \quad k \neq i \quad \vee \quad \forall j.
\end{split}
\end{equation}
This formulation uses the so-called ``big-M method''~\cite{minlp_book}, where a large constant $M$ and a binary variable $\delta_{i,k}$ are introduced, to express the non-overlapping constraints in a general way with respect to the task ordering and allocation. 
If the same agent does not execute both tasks, then $2-a^j_i-a^j_k \neq 0$ and~\eqref{s_i_and_s_k} always hold. In contrast, if $a^j_i=a^j_k=1$, the constraints are managed by $\delta_{i,k} $. If $\delta_{i,k}=1$, the first and third constraints are active ($\tau_i$ after $\tau_k$), while the second and fourth always hold and vice versa if $\delta_{i,k}=0$ ($\tau_i$ before $\tau_k$).

\subsection{Performance Constraints}

$\mathcal{C}_{\mathrm{performance}}$ determines the expected task end times based on the agent's performance and the impact of tasks executed in parallel. The formalization of such a constraint is crucial for the paper's contribution and needs some preliminary definitions, which are reported in the next paragraph. Then, two paragraphs will report two different models for this constraint. 

\subsubsection{Synergy among tasks}

\begin{figure}[t]
 \includegraphics[trim={0 0 0 0cm},clip,width=0.9\columnwidth]{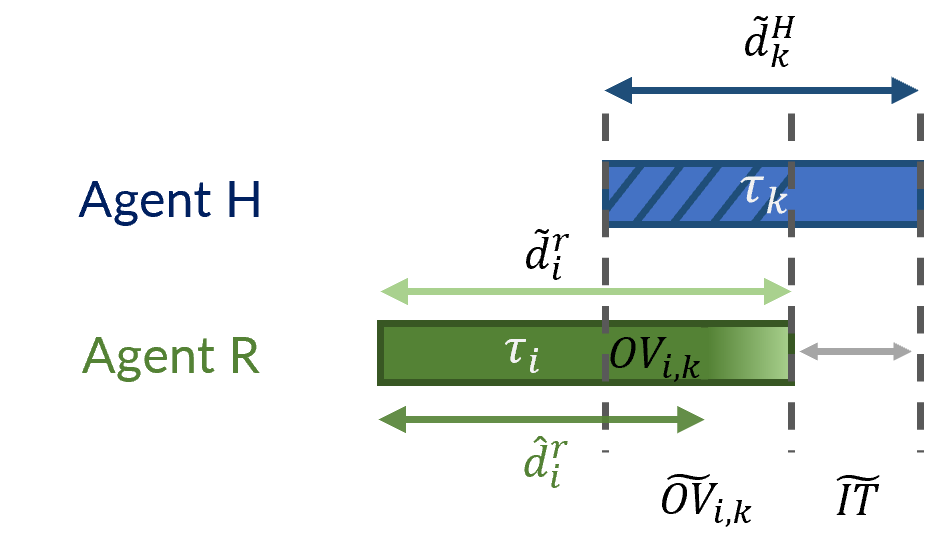}
 \caption{Visualization of the involved variables during the overlapping between human and robot tasks. The measured durations ($\tilde{d}^H_k$ for the human task and $\tilde{d}^r_i$ for a task performed by a generic robot) together with the nominal duration $\hat{d}^r_i$ of the task $\tau_i$ and the nominal $\mathrm{OV}_{i,k}$ and actual $\widetilde{\mathrm{OV}}_{i,k}$ overlapping are reported. }\label{fig: plan_visualization}
\end{figure}

The approach presented here for human-robot synergies is based on the linear model we introduced in \cite{sandrini2022learning}, and refer to Figure \ref{fig: plan_visualization} for a visual understanding of the following formulas.

We can then define the synergy term $s^j_{i,k} \in \mathbb{R}^+$ for task $\tau_i$ performed by the agent $j$ while another agent executes task $\tau_k$. 
To properly formulate it, we denote $d^j_{i}$ and $d^j_{i}\,|\, k$ as the expected duration of $i-$th task the $j$-th agent performs without any other concurrent running task, and the duration under the constraint that another agent performs task $\tau_k$. 
Note that if $\tau_i$ is performed by a robot agent and $\tau_k$ by the human agent, the difference between $d^j_{i}\,|\, k$ and $d^j_{i}$ may be significant. The robot may slow down and even halt to preserve safety. Conversely, $d^j_{i}\,|\, k\equiv d^j_{i}$ if $\tau_i$ and $\tau_k$ are both performed by robot agents since we assume that the trajectories are synchronized.

Under such considerations, we denote $s^j_{i,k}$ as the synergy term:
\begin{equation}\label{synergy_term}
 s^j_{i,k}=
 \left\{
 \begin{aligned}
 \cfrac{d^j_{i}\,|\, k}{d^j_i} & &
 \parbox{16em}{if $\tau_i$ can be concurrent to $\tau_k$, and $\tau_k$ is executed by a human agent} \\
 1 & & \parbox{16em}{otherwise.}
 \end{aligned}
 \right.
\end{equation}
The synergy models the coupling effect between pairs of tasks. If $\tau_k$ causes a slowdown in the robot task (\eg, safety module intervention), then $s^j_{i,k} > 1$; while $s^j_{i,k} < 1$ if the parallelism has beneficial effects. The synergy value will be equal to 1 when parallelism cannot occur, or the coupling effect is neutral.

As shown in Figure \ref{fig: plan_visualization}, the synergy does not apply linearly to the whole task but can modify the temporal overlapping of the task. Specifically, we denote $\mathrm{OV}_{i,k}$ as the overlapping time defined as:
\begin{flalign}\label{equ: overlapping}
 \mathrm{OV}_{i,k} =\left\{
 \begin{aligned}
 & 0, & & \text{if } \Delta T_{i,k} < 0 \\
 & \Delta T_{i,k}, & & \mathrm{otherwise}
 \end{aligned} \right.\quad
 \forall i,k \;\vee\; k>i,
\end{flalign}
where:
\begin{equation}\label{delta_time}
 \Delta T_{i,k} = \min{(t^{e}_i,t^{e}_k)} - \max{(t^{s}_i,t^{s}_k)}.
\end{equation}

We can, finally, see the measure of the overlapping term as the nominal one stretched by the synergy term:
\begin{flalign}\label{equ: scaled_overlapping}
    \widetilde{\mathrm{OV}}_{i,k} = s^j_{i,k} \mathrm{OV}_{i,k}
\end{flalign}

\subsubsection{Human-Aware Task Allocation and Scheduling (STP)}\label{sec: expanded_methodology}

Given the definition of the scaled overlapping through synergy \eqref{equ: scaled_overlapping}, we can define the performance constraints  $\mathcal{C}_{\mathrm{performance}}$ which models the expected task end times based on the agent's performance and the impact of tasks executed in parallel. Specifically, it turns in:
\begin{equation}\label{equ: c_performance}
\begin{split}
\mathcal{C}_{\mathrm{performance}} := 
\left\{\begin{array}{ll}
 \ldots\\
 t^e_i = t^s_i + \sum_{j\in \mathcal{A}}{\hat{d}^j_i a^j_i} + \\ \vspace{-6pt} \\
 \quad\quad \sum_{r \in \mathcal{\mathcal{A}^R}} \sum^{m}_{\substack{k=1 \\ k \neq i}} \mathrm{OV}_{i,k}(s^r_{i,k}-1) a^r_i a^H_k \\
 \ldots
 \end{array}\right.\\
 \forall i. 
\end{split}
\end{equation}
The task end time $t^e_i$ is determined by the start time $t^s_i$, plus a first term that sets the nominal duration based on the selected agent and a second term that adjusts this duration based on parallelism with other tasks, considering the estimated synergy factors between parallel agents. Only the portion of the task that overlaps with the parallel task is affected by the coupling effect and is scaled according to the synergy value.

\subsubsection{Relaxed Human-Aware Task Allocation and Scheduling Formulation (R-STP)}\label{sec: relaxed_methodology}

{The introduction of synergy terms $s^r_{i,k}$ into the model involves a set of closely coupled constraints~\eqref{s_i_and_s_k}-\eqref{equ: c_performance}.
Intuitively, this can create a loop effect where changes in overlapping affect the robot's task duration, which in turn alters the overlap again, making the problem computationally demanding.}
{To address this issue, we propose an approximated model to reduce the computational complexity}. 
We simplify the problem by decoupling the synergy from the task duration~\eqref{equ: c_performance} while adding a penalty term to the cost function.
Thus, we replace~\eqref{equ: c_performance} with:
\begin{equation}\label{equ: c_performance_relaxed}
 \mathcal{C}_{\mathrm{performance}}:= \quad t^e_i = t^s_i + \sum_{j\in \mathcal{A}}{\hat{d}^j_i a^j_i} \quad \forall i \in \left\{1,\dots,m\right\}
\end{equation}
and~\eqref{makespan_cost_function} with a proxy for the makespan:
\begin{equation}\label{relaxed_objective}
 \mathcal{M} = \max_{i=1,\dots,m}{(t^e_i)} + \Delta S
\end{equation}

\noindent where:
\begin{equation}\label{equ: delta_s_multi_robot}
 \Delta S = \sum_{r \in \mathcal{\mathcal{A}^R}} \sum^{m}_{i = 1} \sum^{m}_{\substack{k=1 \\ k \neq i}} \mathrm{OV}_{i,k}(s^r_{i,k}-1) a^r_i a^H_k.
\end{equation}

The rationale behind~\eqref{relaxed_objective} is that the makespan is equal to the sum of the nominal duration of the tasks and the lengthening of the various tasks given by the coupling effect between the human and robots' tasks ($\Delta S$). This makes it possible to switch from the nominal makespan (considering there is no effect between one agent and another) to the actual makespan that considers the reciprocal effect between humans and robots.


\section{Learning of human-robot synergies}\label{sec: synergy_estimation}

Constraints \eqref{equ: c_performance} and \eqref{equ: c_performance_relaxed} need an estimate of the synergy term. 
The learning approach for human-robot synergies is based on the linear model we introduced in \cite{sandrini2022learning} that relates the measured duration of human tasks to the level of parallelism of each robot task through the synergy terms defined in~\eqref{synergy_term}. 

Precisely, for each observation collected in the dataset $\mathcal{D}$, we extract the measures of:
\begin{enumerate*}[label=(\roman*)]
  \item the actual duration of each human task $\tau_k$, denoted as $\tilde{d}^H_k$;
  \item the actual overlapping time between each pair of tasks $\tau_k$ and $\tau_i$, denoted as $\widetilde{\mathrm{OV}}_{i,k}$;
  \item the actual idle time of the task $\tau_k$, denoted as $\widetilde{\mathrm{IT}}^{H}_{k}$.
\end{enumerate*}
Given these measures, the following model holds for a generic pair of robot ($r$) and human:
\begin{equation}\label{equ:time_model}
\begin{split}
\tilde{d}^{H}_k\, = \,
    \sum_{\substack{i=1 \\ i \neq k}}^{m} \Biggl(\, \widetilde{\mathrm{IT}}^{H}_{k} + \widetilde{\mathrm{OV}}_{i,k} \Biggl)\, a_i^r, \quad \forall r \in \mathcal{A}^R.    
\end{split}
\end{equation}
This formulation is made only for the pair of human-robot agents as we are interested in the duration variation caused by the human operator on the robots' tasks.


Synergy coefficient values are estimated in terms of distribution probability following the Bayesian approach. To do so, we assume that the measurements $\tilde{d}^H_k$ follow a normal distribution around the model in \eqref{equ:time_model}, and using \eqref{equ: scaled_overlapping} that relates the actual overlapping with the nominal overlapping given the generated plan through the synergies coefficients, we get:

\begin{equation}\label{task_duration_model}
 d^{H}_k   \sim \mathcal{N}\Biggl(\,
 \sum_{\substack{i=1 \\ i \neq k}}^{m} 
 \left(\, \widetilde{\mathrm{IT}}^{H}_{k} + s^r_{i,k}\, {\mathrm{OV}}_{i,k} \right)a^r_i
 \, , \,
 \sigma_m^{2} \Biggl).
\end{equation}
To capture the uncertainty of the proposed model, we assume that the standard deviation $\sigma_m$ follows a uniform distribution with lower bound $l_b$ and upper bound $u_b$:
\begin{equation}\label{sigma_model}
 \sigma_m \sim \mathcal{U}(l_b,u_b).
\end{equation}
The prior knowledge of the synergy terms, defined as ratios relative to the average duration of a task, inherently holds non-negative values. 
Hence, it is reasonable to assume a log-normal distribution, with $\mu_s$ and $\sigma_s$ respectively, the mean and the standard deviation of the log of the distribution: 
\begin{equation}\label{synergy_model}
 s^r_{i,k} \sim LogN(\mu_s,\sigma^2_s).
\end{equation}

Given the above prior knowledge of the model, the posterior distribution of the estimated synergy parameters $S^r_{k}=(s^r_{1,k},\dots,s^r_{m,k})$ can be obtained through the application of Bayes' Theorem:
\begin{equation}\label{bayes_th}
 p(S^r_k|\mathcal{D}) = \frac{p(\mathcal{D}|S^r_k)p(S^r_k)}{p(\mathcal{D})}.
\end{equation}
The posterior distribution 
is computed using a Markov-Chain Monte Carlo (MCMC) algorithm, such as No-U Turn Sampler (NUTS)~\cite{hoffman2014no} that gives an unbiased estimate on the limit. It is possible to update the prior knowledge distributions using measurements of task durations for both the person and the robot, along with the percentage of overlap. This can be done either after a batch realization of the tasks or after each execution. The method allows for progressive estimates of the probability distributions of human-robot synergies, making the model robust to stochastic human behavior and applicable in real-world cases.

\section{Simulations}

\begin{figure}[t]
 \centering
 \includegraphics[trim={0 0 0 0cm},clip,width=0.9\columnwidth]{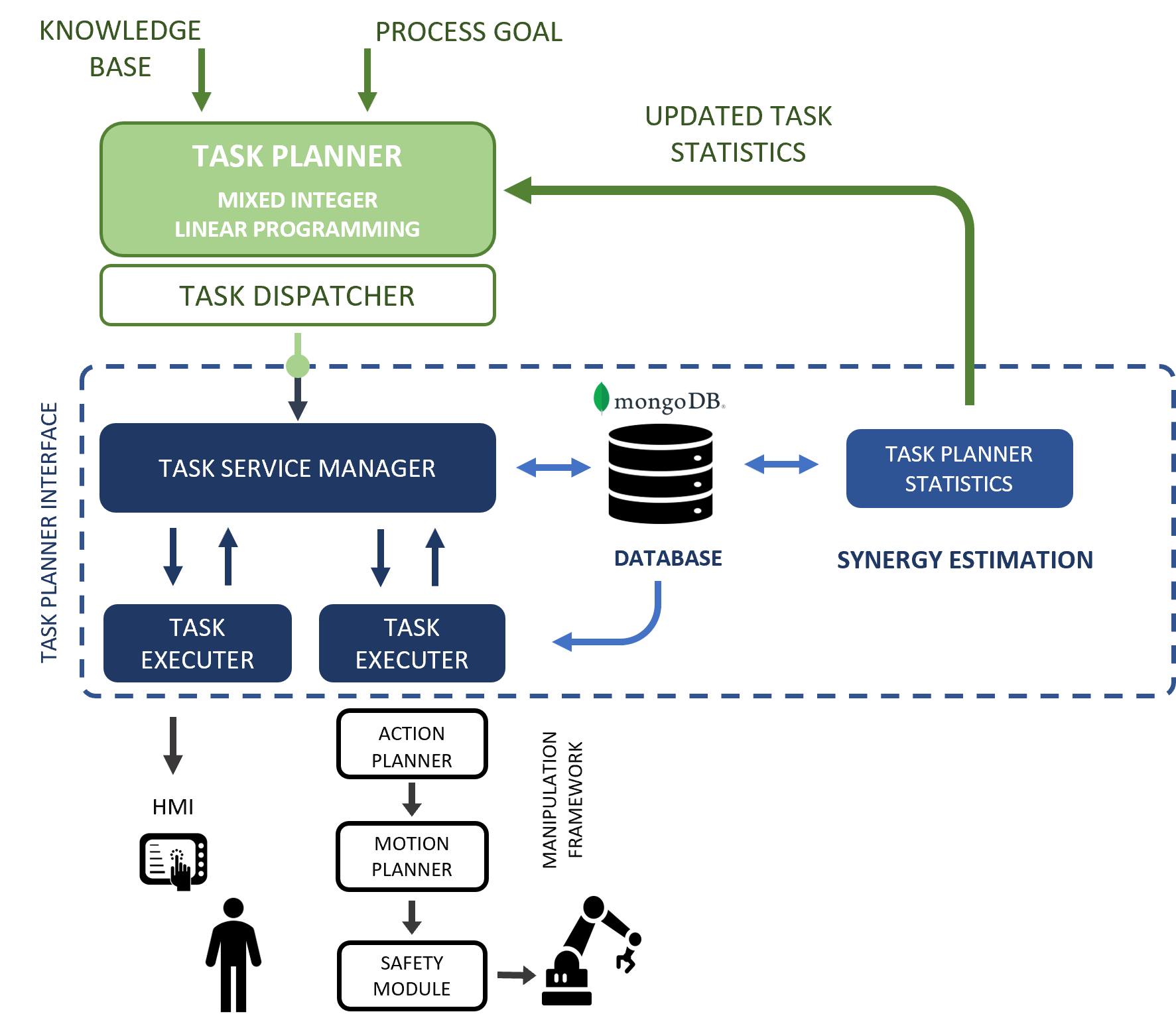}
 \caption{Framework architecture: a three-tiered structure that manages the pipeline from planning to execution.}\label{fig:framework_architecture}
\includegraphics[trim={2cm 0 5cm 0cm},clip,width=0.9\columnwidth]{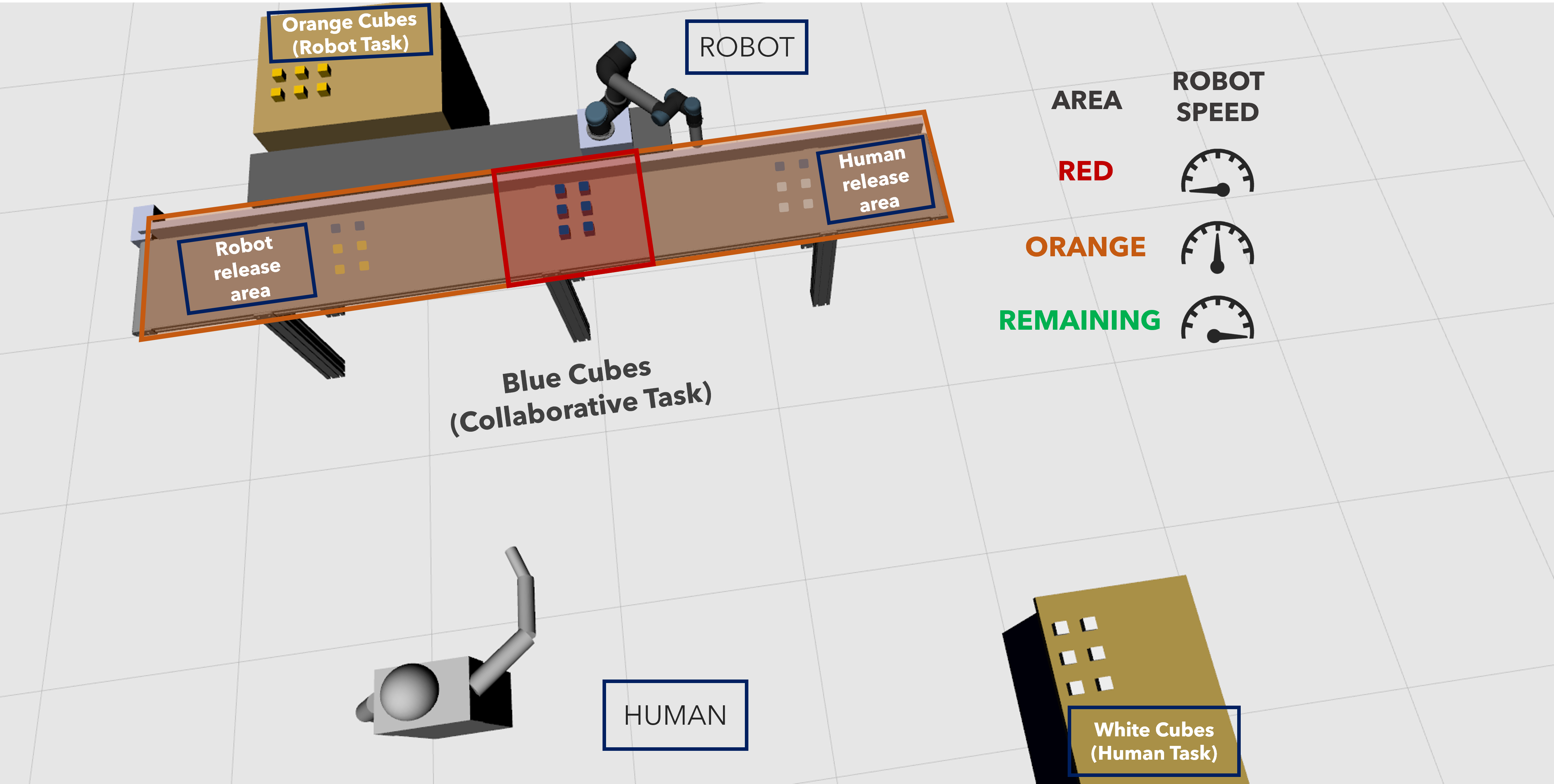}
\caption{Simulated setup. The robot workspace contains orange boxes, the human's white boxes, and the agent's release areas. Blue boxes are shared between agents. Highlighted are the safety areas used in scenario 1.}\label{fig:simulation_setup}
\end{figure}

\subsection{Framework Architecture}\label{sec: framework}

The software architecture is derived from~\cite{sandrini2022learning} and is improved here in the planning, dispatching, and statistics modules (see Figure~\ref{fig:framework_architecture}). The architecture is designed to manage the pipeline from planning to execution. 
At the highest level, \emph{task planning module} solves the allocation and scheduling problem. The \emph{task dispatching module} (Algorithm~\ref{alg:task_dispatcher}) handles the task execution request to the lower levels based on the obtained task plan.
At the lowest level, there is a \emph{task execution module} for each agent.
The task executor grounds symbolic tasks into geometric targets, solves the motion planning problems, and monitors the execution.
In simulation, the human agent is managed like a robotic agent, while in a real-world case study, the user sends the execution request to an HMI and waits for feedback.

The \emph{task planner interface} is between the higher and lower levels.
This module receives task execution requests from the dispatcher, interacts with a MongoDB database~\cite{Mongo_VS_MySql}, and stores information such as task start and completion time.
The \emph{task planner statistics module} is responsible for keeping up-to-date task execution statistics such as the expected duration and performing the synergy estimation between task pairs.

The MINLP models presented in Section~\ref{sec: methodology} are integrated at the task planning level and implemented using GurobiPy~\cite{gurobi}. In addition, the task dispatcher is implemented using the Algorithm~\ref{alg:task_dispatcher} and exploits ROS-topic communication for task requests/responses. The Bayesian estimation of synergy terms proposed in Section~\ref{sec: synergy_estimation} is implemented using Pyro~\cite{bingham2019pyro} and integrated as a ROS-Service in the statistics module.
\begin{algorithm}[t]
 \caption{Task Dispatcher Algorithm}\label{alg:task_dispatcher}
 \begin{algorithmic}
 \Require$\pi = \left\{(t^s_i,t^e_i,a^j_i)\right\} \quad \forall i \in \left\{1,\dots,m\right\}, j \in \mathcal{A}$
 \State\ Sort $\pi$ based on $t^s_i$
 \While{$\neg\,$ $isEmpty(\pi)$}

 \For{$j \in \mathcal{A}$ }

 \State\ $\pi^{j}$ is the tasks in $\pi$ for which $a^j_k=1 \; \forall k$-tasks

 \If{$isEmpty(\pi^{j})$}
 \State\ continue
 \EndIf\

 \State\ $\tau_i \gets getFirstAction(\pi^{j})$
 \State\ $t \gets getTime()$ \Comment{Get current simulation time}

 \If{$t \geq t^s_i $ \& $isFree({Agent}^j)$}

 \State\ Send the execution request of task $\tau_i$ to agent $j$
 \State\ $\pi.pop(\tau_i)$
 \State\ $setBusy({Agent}^j)$
 \EndIf\
 \EndFor\

 \EndWhile
 \end{algorithmic}
\end{algorithm}

\subsection{Simulation Setup}\label{sec: simulation}

The simulated HRC application involves pick-and-place operations for the composition of a mosaic. The application employs a UR5 collaborative robot mounted on a linear axis, working alongside a human operator. Each agent has its workspace where their respective boxes are placed: white boxes for the robot and orange boxes for the operator. The shared workspace contains blue cubes that both agents can manipulate.

The application was tested in two scenarios based on safety specifications aligned with ISO/TS 15066~\cite{ISOTS15066}.

The safety areas of scenario 1 (S1) are defined to comply with the \emph{Speed And Separation Monitoring (SSM)} described in~\cite{ISOTS15066}. In Figure~\ref{fig:simulation_setup}, the safety areas of the proposed application are highlighted. When the human operator enters the orange area, the robot operates at 50 \% of its nominal speed. If the person enters the red area, the robot stops. In the remaining area, the robot moves at its nominal speed~\cite{SafetyAreas}. 
{This safety implementation is typical of industrial applications as it can be achieved with safety-rated sensors (\eg, radar or laser barriers).} 
The task-level goal for each agent is to compose a mosaic in their own releasing area using four proprietary boxes (white for the robot and orange for the human) and two shared boxes (blue ones) for a total of 24 tasks (12 picking and 12 placing). 

Scenario 2 (S2) is based on the robot's velocity modulation to prevent it from colliding with a human. By continuously monitoring the minimum agent's distance, it is possible to calculate the maximum velocity of the robot toward the human ($v_{\mathrm{hr_{max}}}$) following the definition of the minimum separation distance provided by the \emph{SSM} (refer to~\cite{ISOTS15066}):
\begin{equation}\label{velocity_hr_disequality}
 v_{\mathrm{hr_{max}}}(t) \geq \sqrt{ v^2_h+(a_s T_r)^2 - 2 a_s(C-S_d(t)) }-a_s T_r-v_h
\end{equation}
in which the involved parameters are detailed in Table~\ref{table: SSM_parameters}.
\begin{table}[t]
 \centering
 \renewcommand{\arraystretch}{1.05} 
 \caption{SSM parameters description. ISO/TS 15066 specifies human's velocity $v_h$ when not tracked. Robot and vision system reaction times are cumulated.}\label{table: SSM_parameters}
 \centering
 \begin{tabularx}{\columnwidth}{ccX}
 \hline
 \textbf{Symbol} & \textbf{Value} & \textbf{Description} \\
 \hline
 $v_h$ & \SI{1.6}{\metre\per\second} & Human velocity toward the robot\tnote{*} \\ 
 $a_s$ & \SI{1}{\metre\per\second\squared} & Max. robot deceleration \\
 $T_r$ & \SI{0.3}{\second} & Max. system reaction time\tnote{**} \\
 $C$ & \SI{0.2}{\metre} & Position uncertainty \\
 $S_d$ & Measured & Separation distance \\
 \hline
 \end{tabularx}
\end{table}

The task-level goal is for each agent to compose a mosaic in their own releasing area using four proprietary boxes for the robot (orange boxes), three proprietary boxes for the human operator (white boxes), and two shared cubes (blue ones).


In the first phase of simulations, 50 random plans were run to estimate the average duration of individual tasks and synergies.
Once estimates were given, task planning was run.
Regarding the parameters in Section~\ref{sec: synergy_estimation} for synergy estimation, we ensured the median value of 1 of the synergy terms and $\sigma_s = 0.5$.


\subsection{Baseline and Metrics}\label{sec: baseline}

We compare our methods with two baselines:
\begin{enumerate*}[label=(\roman*)]
 \item \emph{Baseline TP}, which is a simplified version of~\eqref{milp_optimization_problem}, and also similar to~\cite{sandrini2022learning} (the coupling effect between tasks pairs is not taken into account);
 \item the MILP model proposed in~\cite{lippi2021mixed}, where spatially close tasks are constrained not to be executed in parallel by the agents (referred to as \emph{Not Neighboring TP}).
\end{enumerate*}

The evaluation of the proposed method against different task planners focused on performance and safety.
For performance evaluation, we measure the plan execution duration as follows:
\begin{equation}\label{eq:makespan-eval}
 \mathrm{Makespan} = \max_{i \in \left\{ 1, \, \dots \, , \, m \right\}}{(\tilde{t}^e_i)}
\end{equation}
where $\tilde{t}^e_i$ is the measured end time of a task $\tau_i$ in a task set $\mathcal{T}$ during the execution of a plan $\pi$.
As for safety, we monitor with the vision system the minimum distance ($D_{min}$) between the human operator and the robot during the execution of the plans.
Then, we compute the cumulative probability distribution of such distance over all the executions of each method as:
\begin{equation}\label{average_cumulative_probability_metric}
 \frac{1}{N_{\mathrm{plans}}}\sum_{k=1}^{N_\mathrm{plans}} {\mathbb{P} \bigl(D_{\min} \leq \tilde{d} \bigl) \qquad \forall \tilde{d} \in [0, d_{\max}]}
\end{equation}
Thus, this index represents the probability that $D_{\mathrm{min}}$ falls below a given distance $\tilde{d}$ (between 0 and the maximum distance $d_{max}$ of the vision system) for the method at hand.

\subsection{Results}\label{sec: results}

First, we evaluate the estimated synergies for both scenarios.

\begin{figure*}[h!]
 \centering
 \begin{minipage}{0.49\textwidth}
     \centering
\subfloat[][Estimated synergy  terms (robot side) in matrix format]{\includegraphics[width=0.8\columnwidth]{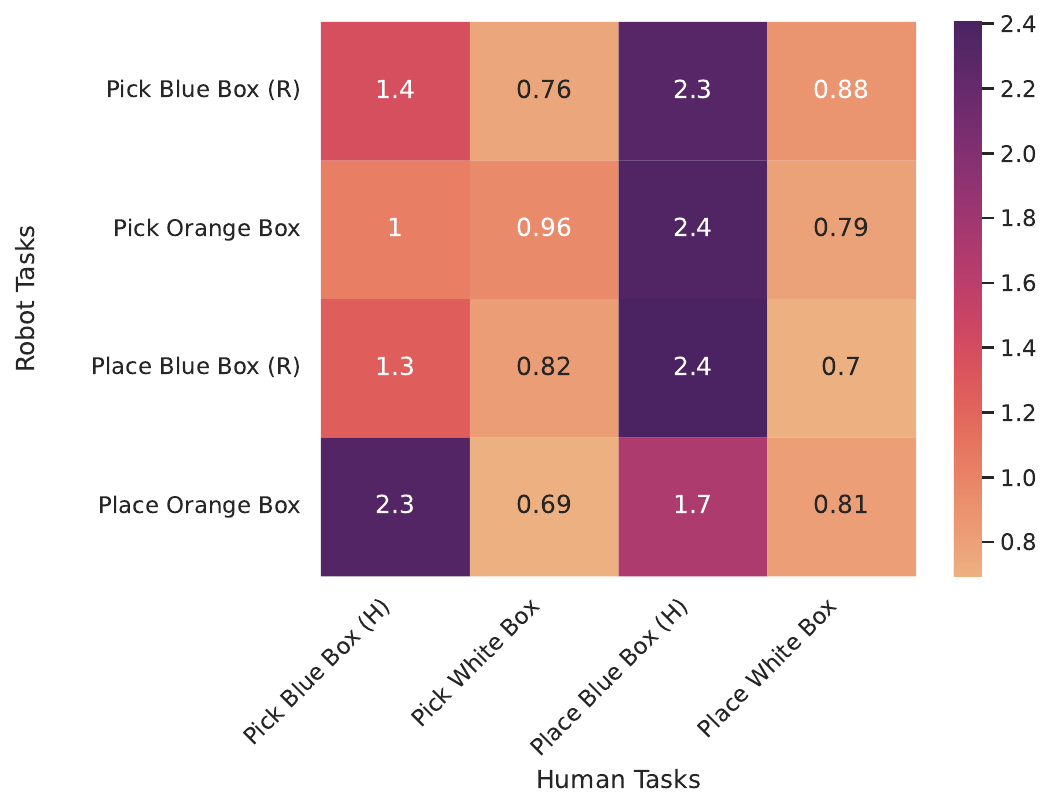}\label{}\label{fig:synergy_matrix_safety_areas}}  \\
\subfloat[][Plan execution duration]{\includegraphics[width=.8\columnwidth]{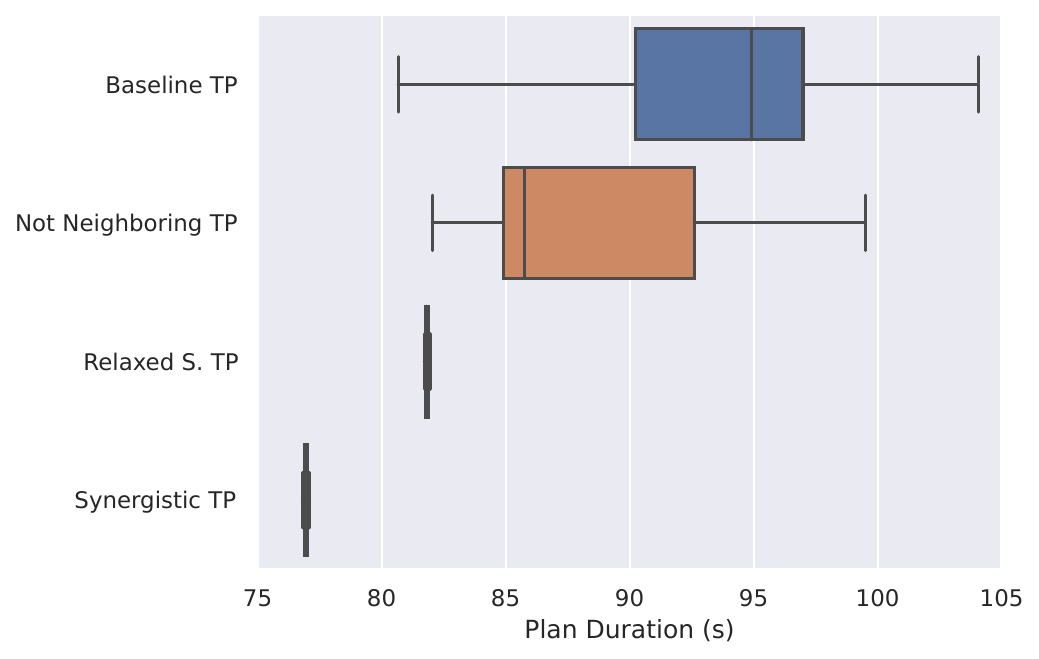}\label{fig:duration_comparison_scenario_1}} \\
 \subfloat[][Cumulative probability of the minimum human-robot distance]{\includegraphics[width=0.8\columnwidth]{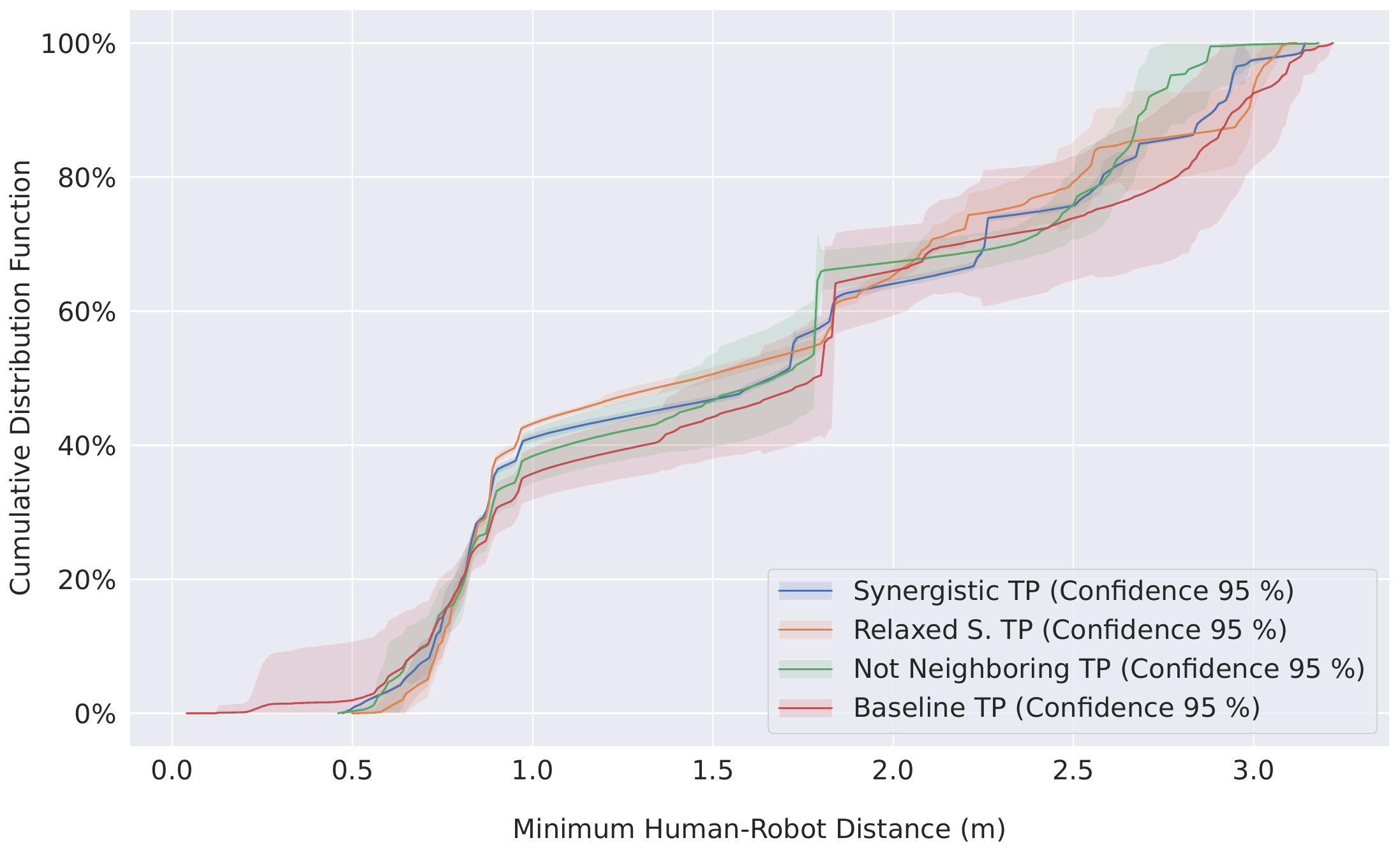}\label{fig:distance_comparison_scenario_1}}
 \caption{Results of \emph{Scenario 1: static safety zones}}\label{fig:scenario_1}
 \end{minipage}
 \begin{minipage}{0.49\textwidth}
 \centering
\subfloat[][Estimated synergy  terms (robot side) in matrix format]{\includegraphics[width=.8\columnwidth]{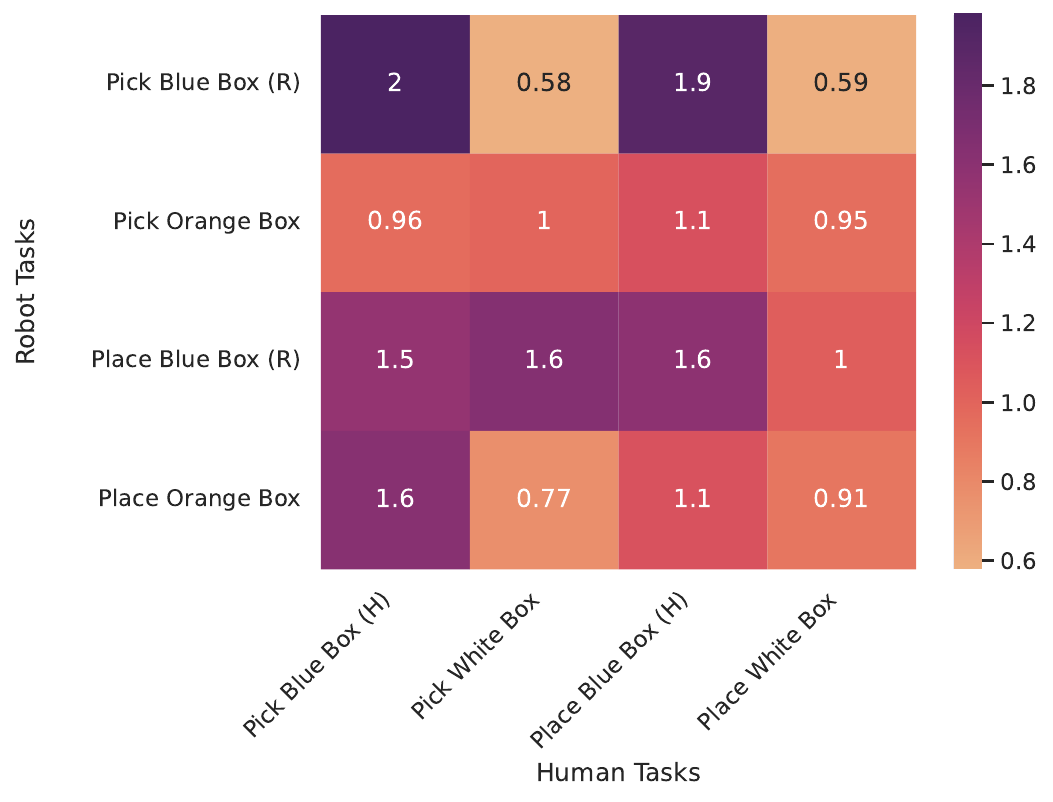}\label{fig:synergy_matrix_safety_scaling}}\\
\subfloat[][Plan execution duration]{\includegraphics[width=.8\columnwidth]{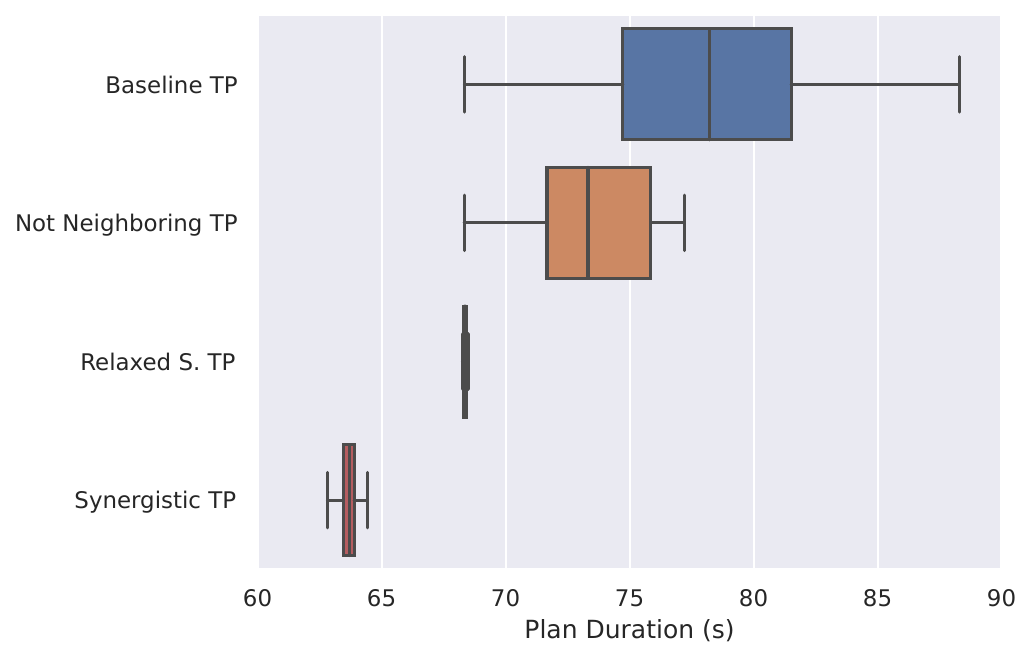}\label{fig:duration_comparison_scenario_2}}\\
 \subfloat[][Cumulative probability of the minimum human-robot distance]{\includegraphics[width=.8\columnwidth]{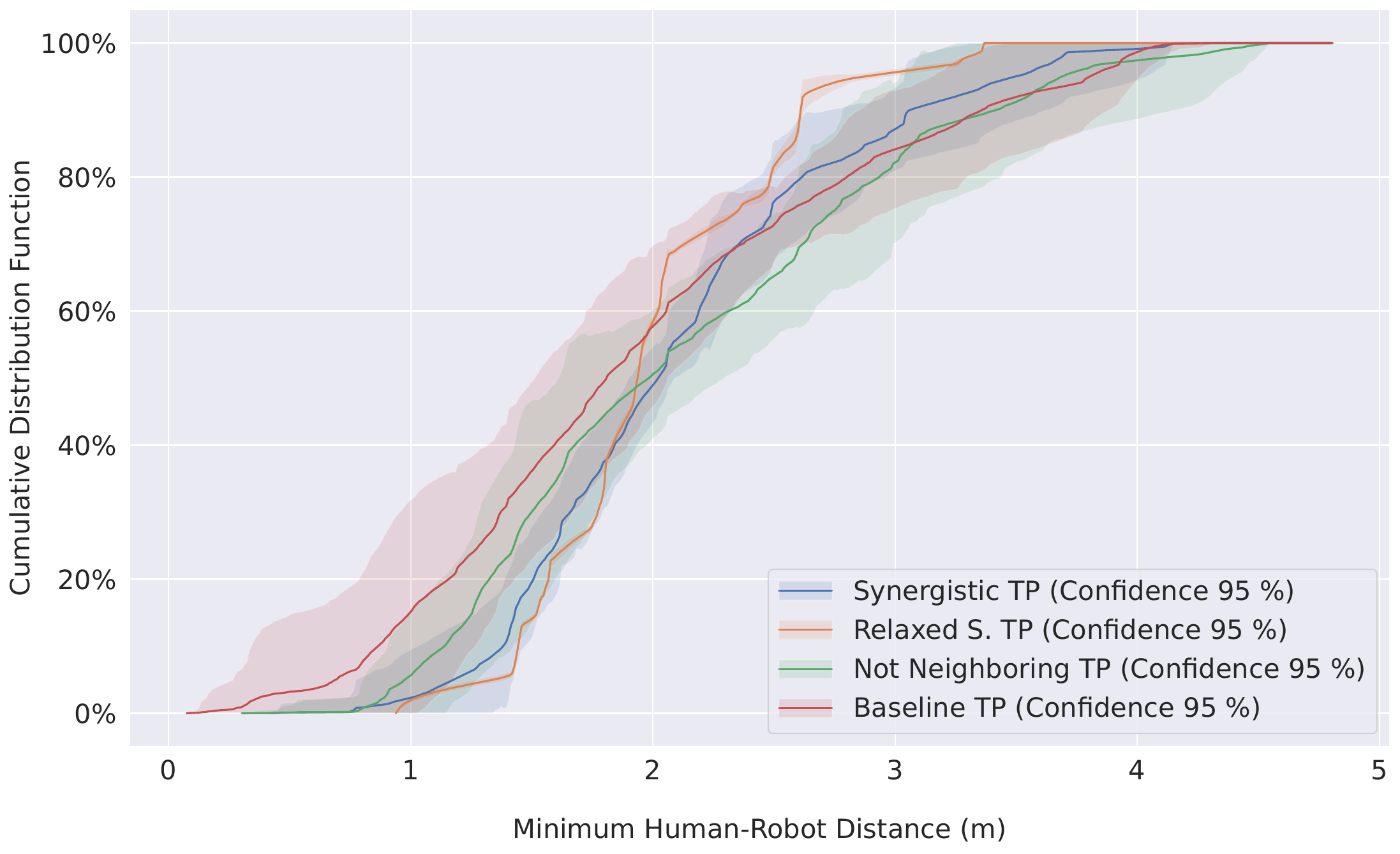}\label{fig:distance_comparison_scenario_2}}
 \caption{Results of \emph{Scenario 2: continuous safety velocity scaling}}\label{fig:scenario_2}
 \end{minipage}
\end{figure*}

The heat maps in Figures~\ref{fig:synergy_matrix_safety_areas} and \ref{fig:synergy_matrix_safety_scaling} show the synergy for each robot task (matrix rows) and for each concurrent human task (matrix columns).
A synergy greater than one means that the robot task takes more time than its average duration (thus, the coupling penalizes the plan execution).
An index smaller than one implies that the task takes less time than its average (leading to a more efficient plan).
For Scenario 1 (Figure~\ref{fig:synergy_matrix_safety_areas}), the most significant slowdown occurs when the human operator places the shared boxes (blue boxes) in their release area.
The start point causes the robot to stop at the execution start since the static point selection impacts the whole agent's trajectory.

The coupling effects between tasks also arise in Scenario 2, where the scaling factor acts more smoothly, changing continuously based on~\eqref{velocity_hr_disequality}. The results of the synergy estimation of this scenario are reported in Figure~\ref{fig:synergy_matrix_safety_scaling}.
The first row of the heat map shows that the tasks of picking/positioning shared boxes (blue boxes) by humans cause similar slowdowns of the robot, unlike the case of parallelism with picking/positioning of humans' boxes (white boxes), which causes faster execution than the average duration.
The robot's task of picking the orange object is unaffected by any human tasks; vice versa, the robot's task of placing the blue box is significantly slowed down, except when the human performs the white box placement.

Table~\ref{table: comparison_optimized_plans} compares the makespan~\eqref{eq:makespan-eval} and the synergy of executed plans.
Figures~\ref{fig:scenario_1} and~\ref{fig:scenario_2} show the results of the online simulations of Scenarios 1 and 2.
Figures~\ref{fig:duration_comparison_scenario_1} and~\ref{fig:duration_comparison_scenario_2} compare each planner's makespan in 50 executions.
In both scenarios, \emph{STP} outperforms the other methods by reducing the makespan of about 
$18\%$ w.r.t.\ Baseline TP and $13\%$ w.r.t.\ Not Neighboring TP.\@ The R-STP shows an average plan execution time comparable with the best case of the Baseline and the Not Neighboring TP;\@ the average makespan obtained with the Relaxed-STP method outperforms the Baseline (reduction of about $13\%$ for both scenarios) and the solutions obtained from Not Neighboring TP (reduction about $7\%$ for both scenarios).

Figures~\ref{fig:distance_comparison_scenario_1} and~\ref{fig:distance_comparison_scenario_2} show the average cumulative probability distribution of the minimum human-robot distance~\eqref{average_cumulative_probability_metric}, measured during the execution of the plan.
In Scenario 1, the STP and Relaxed-STP show a cumulative probability distribution comparable with the Not Neighboring TP method for distances about $\SI{0.8}{\metre}$, with a minimum human-robot distance of about $+\SI{0.4}{\metre}$ greater than that of the baseline case ($\SI{0.04}{\metre}$).
Relaxed-STP has the lowest probability of having samples below $\SI{0.8}{\metre}$.
In Scenario 2, these differences are more evident for larger distances.
Relaxed-STP is the least likely to have distances below $\SI{1.8}{\metre}$, followed by STP.
Overall, Relaxed-STP can be regarded as the safest method, with a minimum distance $+\SI{0.86}{\metre}$ greater than Baseline TP ($\SI{0.07}{\metre}$), followed by STP ($+\SI{0.34}{\metre}$) and Not Neighboring TP with minimum distances greater than about $+\SI{0.23}{\metre}$ than Baseline TP, respectively.

\begin{table*}[t]
 \caption{Comparison of task planners: performance summary of optimized plans. The term $\Delta S$ is computed according to the definition in~\eqref{synergy_term}. The average, minimum, and maximum values are calculated over all $N_{\mathrm{plans}}$. Scenario 1 with the safety areas is referred as S1, while Scenario 2 with continuous safety velocity scaling as S2.}\label{table: comparison_optimized_plans}
 \centering
 \begin{tabularx}{\textwidth}{XGLGLGLGL}
 \toprule
 \textbf{Method} &
 \multicolumn{2}{c}{\textbf{Makespan (s)}} &
 \multicolumn{2}{c}{\textbf{Average $\Delta S$ (s)\tnote{*}}} &
 \multicolumn{2}{c}{\textbf{Min $\Delta S$ (s)\tnote{*}}} &
 \multicolumn{2}{c}{\textbf{Max $\Delta S$ (s)\tnote{*}}} \\
 & S1 & S2 & S1 & S2 & S1 & S2 & S1 & S2 \\
 \midrule
 Baseline TP & 81.80 & 68.34 & \num{8.90 \pm 7.80} (95\%) & \num{5.05 \pm 13.67} (95\%) & 0.16 & -6.63 & 14.96 & 16.37 \\
 Not Neighboring TP & 81.80 & 68.34 & \num{1.93 \pm 6.00} (95\%) & \num{-4.01 \pm 3.75} (95\%) & -4.00 & -7.20 & 9.66 & -1.56 \\
 Relaxed-STP & 81.80 & 68.34 & -4.65431 & -8.76 & -4.65 & -8.76 & -4.65 & -8.76 \\
 Synergistic TP & 79.61 & 62.11 & -3.94048 & -6.29 & -3.94 & 6.29 & -3.94 & -6.29 \\
 \bottomrule
 \end{tabularx}
\end{table*}

\subsection{
{
Discussion on Computational Complexity}
}
Although this paper is not focused on providing real-time solutions, we discuss the proposed model's computational complexity. {Indeed, the time frame for obtaining a solution must align with the industrial scenario in which the planner computes a solution, and the resulting plan is subsequently put into execution.} The MINLP is generally $\mathcal{NP}$-hard, and the curse of dimensionality of integer variables (\eg, increasing the number of tasks and agents) and handling nonlinearities can lead to suboptimal solutions with limited computation.
For real-world problems such as those presented in the article, the solution consists of tuning the two parameters of the solver.
The first is the MIP optimality gap, which serves as an indicator of solution quality.
It is the relative ratio between the best-known bound on the objective function and the incumbent solution's objective value.
This value is equal to 0 when {is guaranteed that the solution is optimal.}
The second parameter is the maximum available time, depending on the available computation. 

Table~\ref{table: optimization_parameters} shows the parameters used for the experiments and the solution time achieved within the two simulated scenarios {using a solver Gurobi~\cite{gurobi} running on a notebook equipped with an Intel i7-1165G7 2.8 GHz 4-core CPU and 16 GB of RAM. The number of tasks involved in the two scenarios is about 20 and is compatible with use cases of industrial interest and other research work in the field of planning for HRC. Moreover, the achieved computation time is compatible with the ``offline'' planning scenario. If the number of tasks and agents increases significantly, is possible to use hierarchical planning~\cite{hierachical_planning_survey} (\ie, group atomic actions into more complex tasks) or use heuristic methods to speed up the search for approximate solutions~\cite{MILP_heuristic}.}  

\begin{table}[h]
\renewcommand{\arraystretch}{1.3}
\centering
\caption{Optimization Parameters of proposed methods. The timeout time is set to \SI{60}{\second} in both Relaxed-STP, while it is set to \SI{240}{\second} in both Synergistic TP.}
\label{table: optimization_parameters}
\begin{tabularx}{\columnwidth}{XGLGL}
    \hline
    \textbf{Method} & \multicolumn{2}{c}{\textbf{MIP-GAP}} & \multicolumn{2}{c}{\textbf{Solution Time}} \\
     & S1 & S2 & S1 & S2 \\
    \hline
    Relaxed-STP   & \SI{1.70}{\percent} & \SI{2}{\percent} & \SI{18}{\second} &  \SI{42}{\second}\\
    Synergistic TP& \SI{8.}{\percent} & \SI{10.8}{\percent}  & \SI{192}{\second} & \SI{240}{\second}\\
    \hline
\end{tabularx}
\end{table}

\section{Case Study}\label{sec: case_study}

\begin{figure*}[t]
\centering
\begin{minipage}[b]{0.44\textwidth}
    \centering
    \hfill
    \subfloat[Experimental setup.]
    {\includegraphics[trim={0cm 0.15cm 0cm 0.15cm},clip,width=\textwidth]{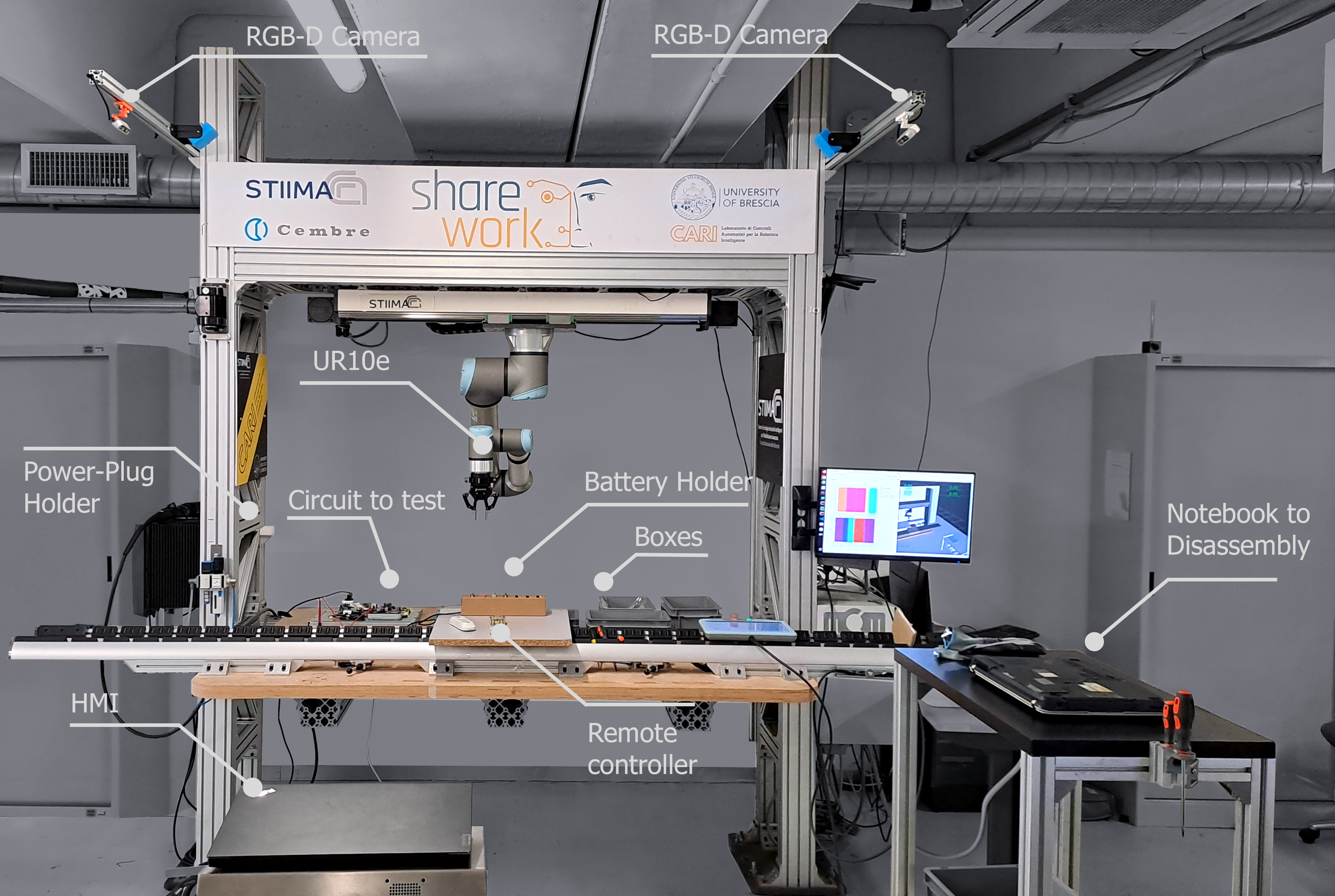}
    \label{fig:real_hardware_setup}}
\end{minipage}%
\begin{minipage}[b]{0.56\textwidth}
    \centering
    \hfill
    \subfloat[Pick\&Place battery]{\includegraphics[width=0.23\textwidth]{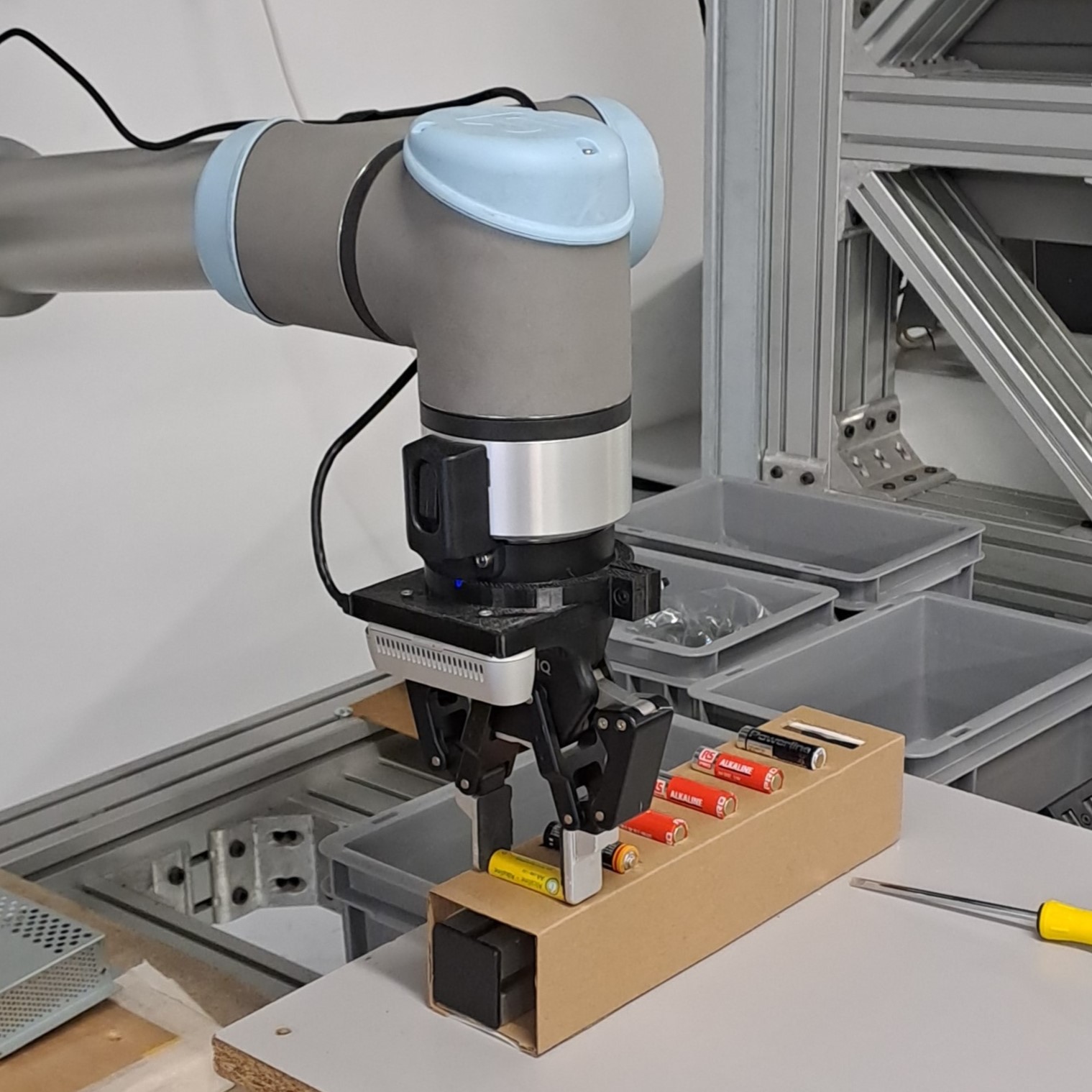}\label{fig:sub1}}
    \hfill
    \subfloat[Board localization]{\includegraphics[width=0.23\textwidth]{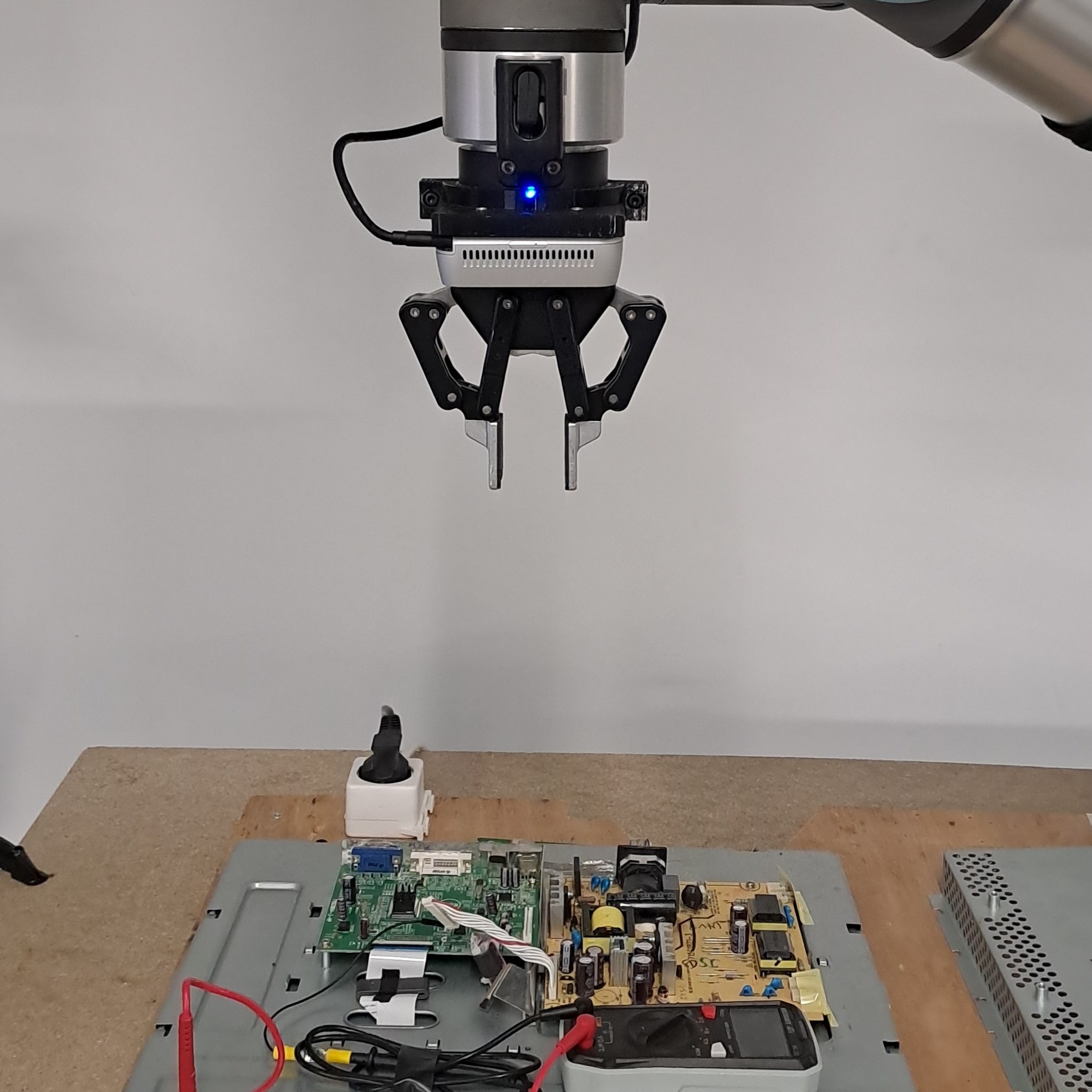}\label{fig:sub2}}
    \hfill
    \subfloat[Remove plug and place it]{\includegraphics[width=0.23\textwidth]{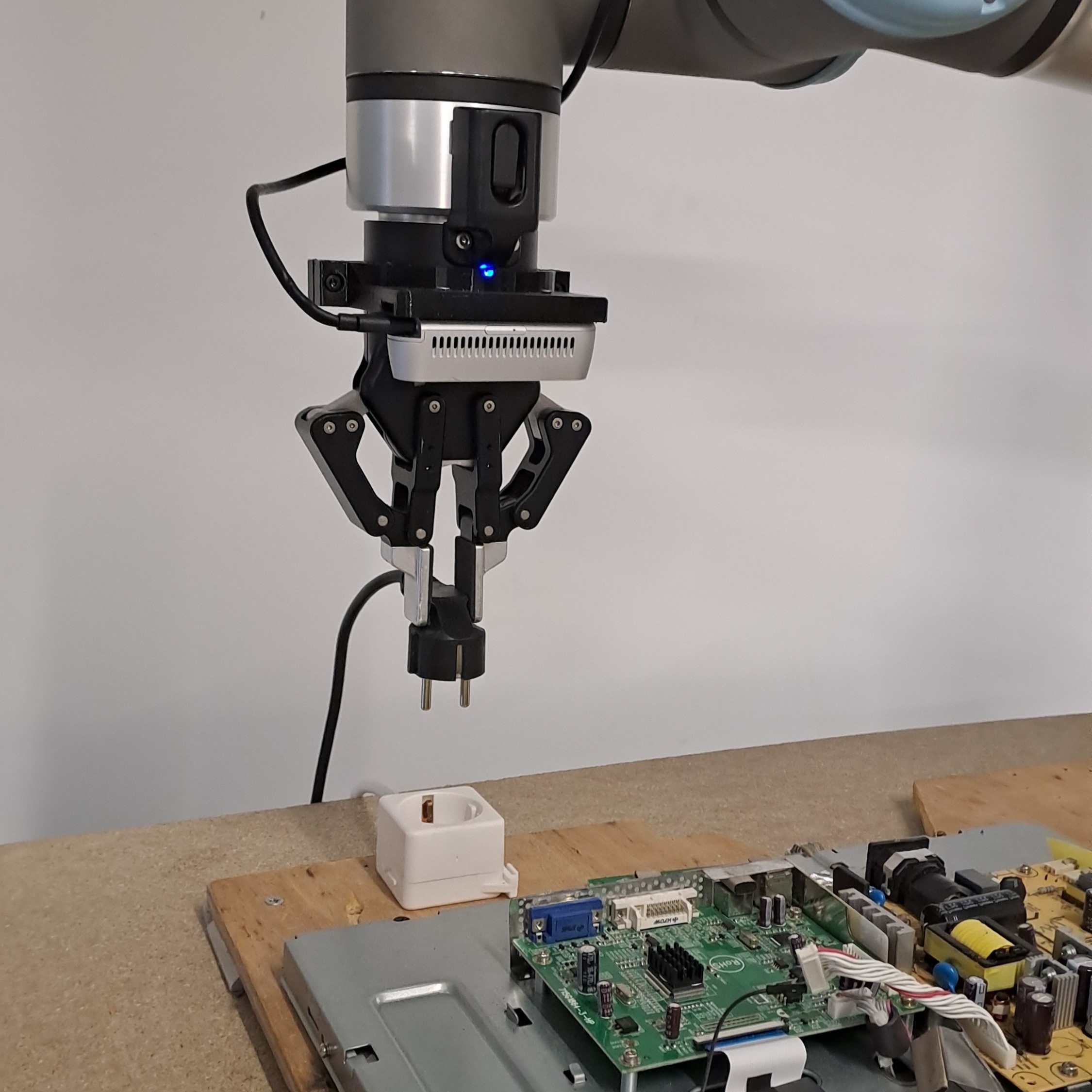}\label{fig:sub3}}
    \hfill
    \subfloat[Probe circuit]{\includegraphics[width=0.23\textwidth]{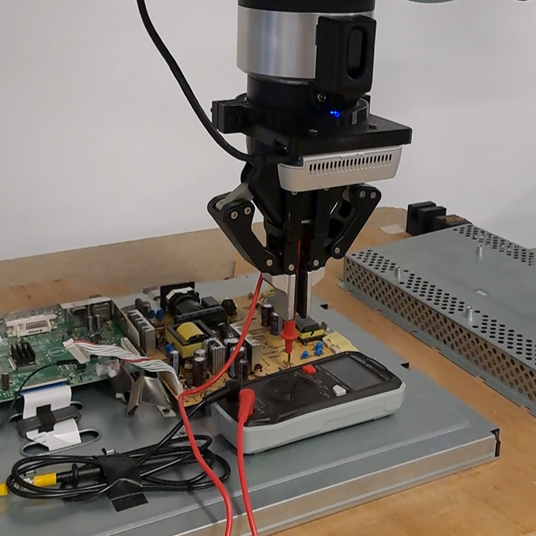}\label{fig:sub4}}
    \\
    \hfill
    \subfloat[Remove notebook cover]{\includegraphics[width=0.23\textwidth]{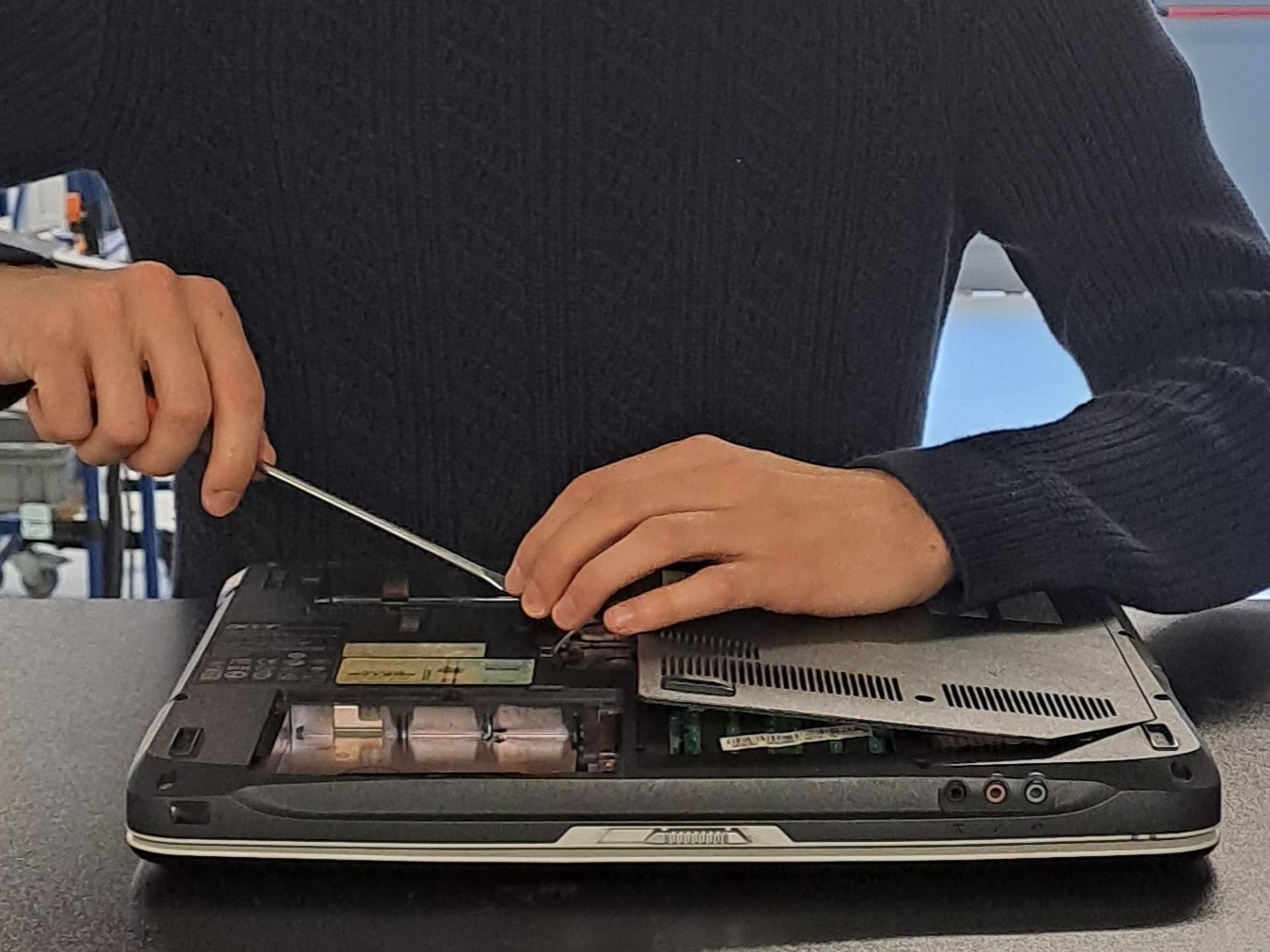}\label{fig:sub5}}
    \hfill
    \subfloat[Remove components]{\includegraphics[width=0.23\textwidth]{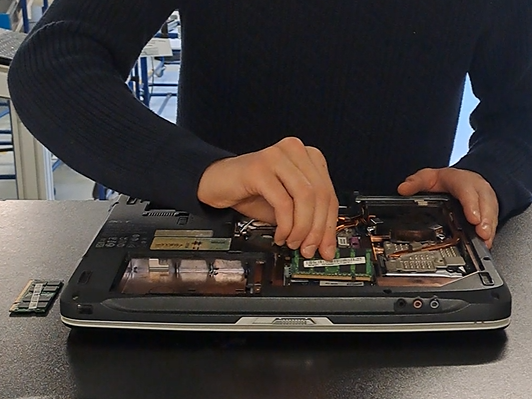}\label{fig:sub6}}
    \hfill
    \subfloat[Disassemble controller]{\includegraphics[width=0.23\textwidth]{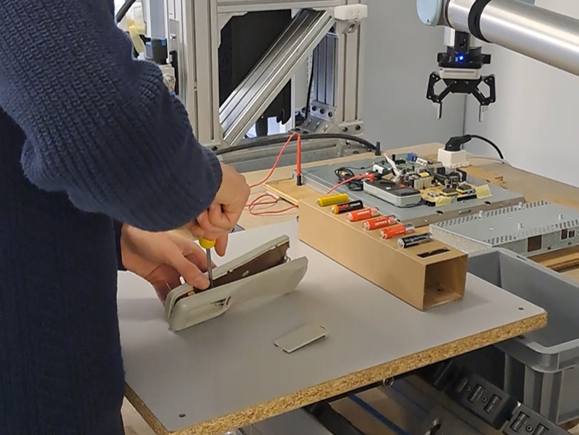}\label{fig:sub7}}
    \hfill
    \subfloat[Simulation environment]{\includegraphics[width=0.23\textwidth]{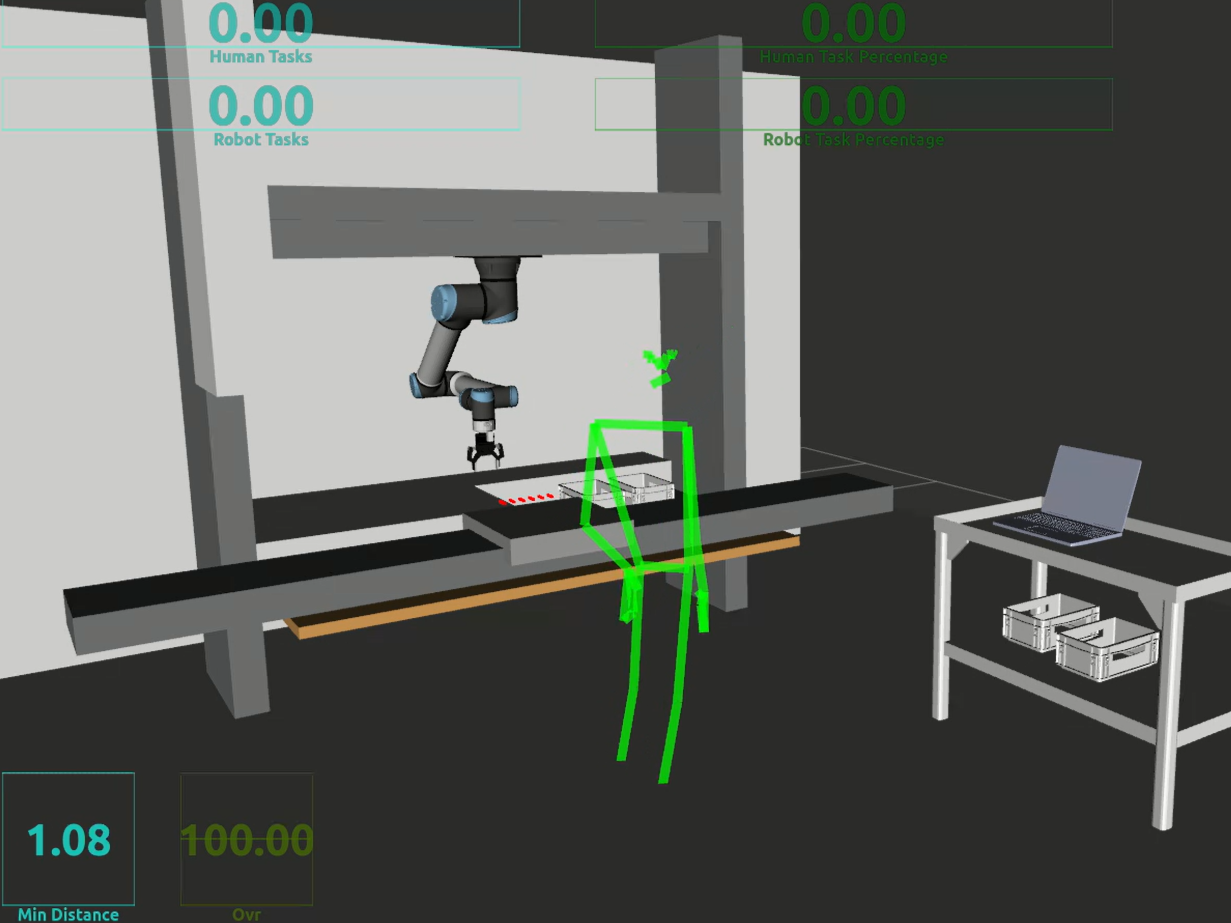}\label{fig:sub8}}
    \hfill
\end{minipage}
\caption{Experimental tests for the disassembly and recycling of e-waste in HRC set. In (a), the experimental setup is shown, (b)-(e) show the robot tasks, (f)-(h) show the human operator tasks, and (i) shows the simulation environment.}
\label{fig:tasks_in_real_case_scenario}
\end{figure*}

We consider an E-waste disassembly and testing case study {inspired to the ShareWork EU project industrial use case~\cite{umbrico2022design}}.
We use the collaborative cell in Figure~\ref{fig:real_hardware_setup}, which consists of a Universal Robots UR10e mounted upside down.
The cell was designed within the EU project ShareWork.
Two Realsense D435 cameras operating at 30 Hz monitor the shared workspace between humans and robots. These cameras provide the human tracking system with RGB-Depth frames.
{Drawing upon established research findings in this field~\cite{scholz2024sensor,palmieri2021human, ferraguti2020safety, ferrari_kalman}, the human pose estimation system uses Mediapipe BlazePose~\cite{bazarevsky2020blazepose} for skeleton tracking on RGB frame, reconstructs the 3D pose with the depth map using the pin-hole model, and applies a Kalman filter (code available~\cite{human_pose_estimation_code}) to reduce noise.}

The experiments are also summarized in the attached video.

\subsection{E-Waste Disassembly and Testing Scenario}
\begin{figure}[h!]
\centering
\subfloat[Estimated synergy matrix.]{
    \includegraphics[trim={0cm 0 0cm 0cm},clip,width=0.8\columnwidth]{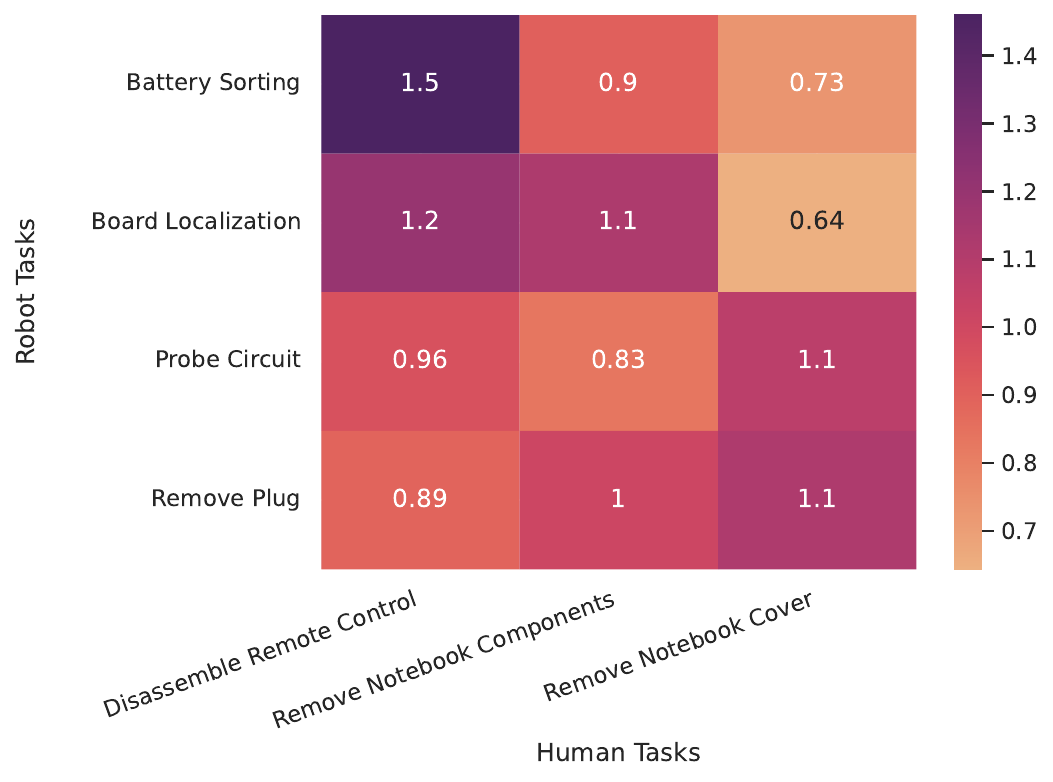}\label{fig:synergy_matrix_real_case}}\\
\subfloat[Plan execution duration.]{
    \includegraphics[trim={0cm 0 0cm 0cm},clip,width=0.8\columnwidth]{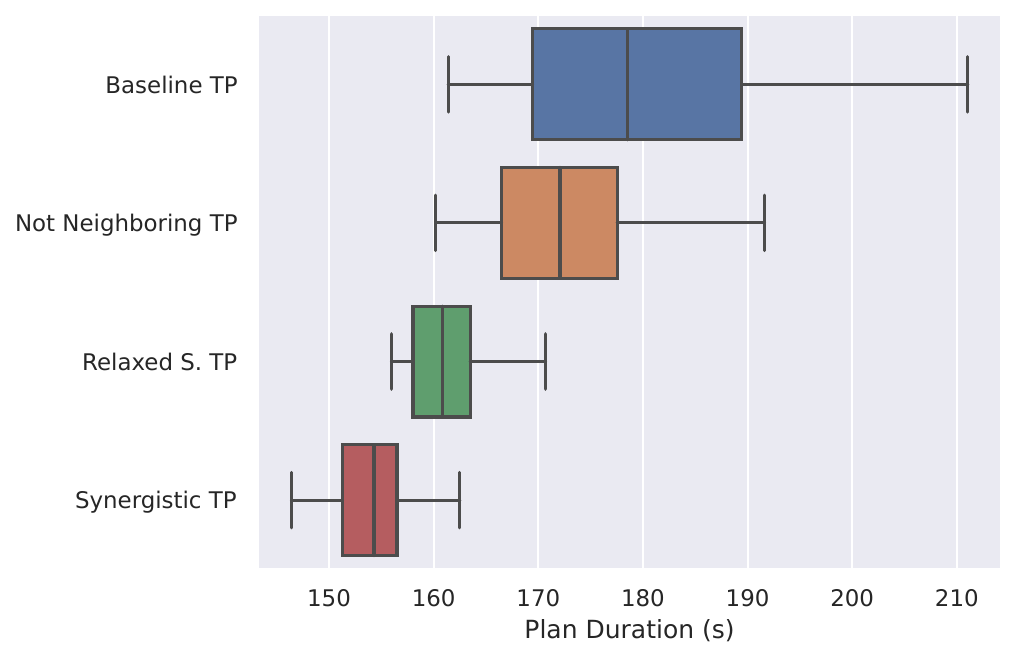}\label{fig:real_case_scenario_plan_duration}}\\
\subfloat[Cumulative probability of the minimum human-robot distance]{
    \includegraphics[trim={0cm 0 0cm 0.89cm},clip,width=0.8\columnwidth]{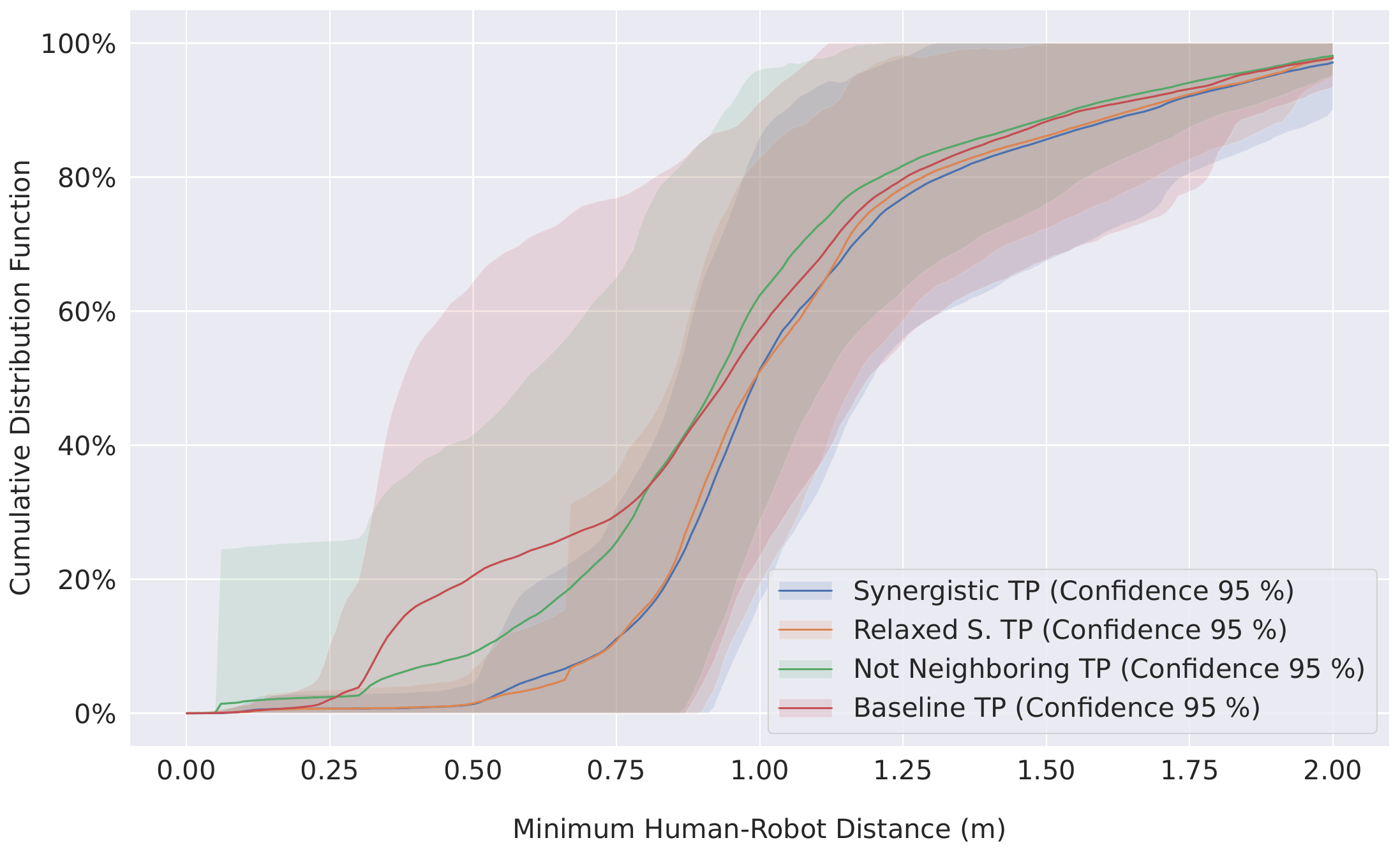}\label{fig:real_case_scenario_distance_comparison}}\\
\captionof{figure}{Results of the real-world use case scenario.}

\end{figure}

The tasks assigned to the agents in the applied scenario are illustrated in Figure~\ref{fig:tasks_in_real_case_scenario}.
The robot is tasked with testing an electrical circuit, which involves locating the board for inspection, probing the circuit with a multimeter probe, disconnecting the power socket, and placing it in a suitable holder. In addition, the robot must place two batteries from the battery holder in an appropriate box (for recovery or disposal). The precedence constraint for the robot is the localization task, which must be completed before any tasks involving the board. Concurrently, a human operator is responsible for disassembling a laptop: first, removing the covers and, subsequently, extracting the components.
He/she must also disassemble two remote controls and store the batteries in the battery holder, from which the robot will retrieve them two times. The remote controller and batteries are placed in a workspace shared with the robot. The total number of tasks is 9; each test takes about 4 minutes.
The process was executed 30 times with random plans (generated by Baseline TP). Figure~\ref{fig:synergy_matrix_real_case} reports the results of such estimation.

\subsection{Experiments Protocol}\label{sec: experiments_protocol_results}

The experiments involved 15 people, including masters, students, PhD students, researchers, and other workers between the ages of 19 and 31.
After initial training on the tasks to execute, the participants performed five plans for each method in randomized order.
While executing the plans, we collected task duration and the human-robot distance.
At the end of each batch of executions, the subjects filled out a questionnaire to assess their subjective evaluation of the cooperation with the robot. 
The questions are in Table~\ref{table: questionnaire_statistical_comparison} and relate to overall satisfaction (Q1), safety (Q2), human-robot distance (Q3), idle time (Q4), perceived fluency assessment (Q5) and robot interruptions (Q6). {The questions are partially inspired by the questionnaire of \cite{lasota2015analyzing} and \cite{petzoldt2022implementation}, adapting and expanding them to suit the specific case under analysis.} {The questionnaire and the related statistical analysis are not meant to be an exhaustive evaluation but aim to provide a preliminary assessment of the workers' perceived impact.  This complements the quantitative evaluation with an initial understanding of the human experience with the proposed method.} The questionnaire follows a five-option response format (from ``strongly disagree'' to ``strongly agree'').

\subsection{Results}
Figure~\ref{fig:real_case_scenario_plan_duration} compares execution times.
Both the proposed methods lead to shorter execution times.
STP (\SI{154.6}{\second} on average) achieved a reduction of \SI{14.7}{\percent} compared to the Baseline TP (\SI{181.2}{\second} on average), and \SI{10.9}{\percent} compared to the Not Neighboring TP (\SI{173.5}{\second} on average).
Similarly, Relaxed-STP (\SI{161.5}{\second}) achieved a reduction of \SI{10.9}{\percent} compared to the Baseline TP and \SI{7.0}{\percent} compared to the Not Neighboring TP.\@

Figure~\ref{fig:real_case_scenario_distance_comparison} compares the human-robot distance measured during the executions; for each human-robot distance, it shows the percentage of samples below that distance measured during the runs.
The results show that the proposed methods are less likely to cause small human-robot distances.
By taking \SI{0.4}{\metre} as an example of risky distance, STP and Relaxed-STP have a percentage of \SI{0.85}{\percent} and \SI{0.90}{\percent} samples below such threshold. In contrast, Not Neighobring TP has \SI{6.75}{\percent} and Baseline TP has \SI{15.95}{\percent}.
It should be noted that the percentage of samples obtained below this distance with the proposed methods is comparable to that of false positives detected by Mediapipe.

\begin{figure*}
 \centering
 \includegraphics[trim={0cm 0.2cm 0cm 0.1cm},clip,width=0.97\textwidth]{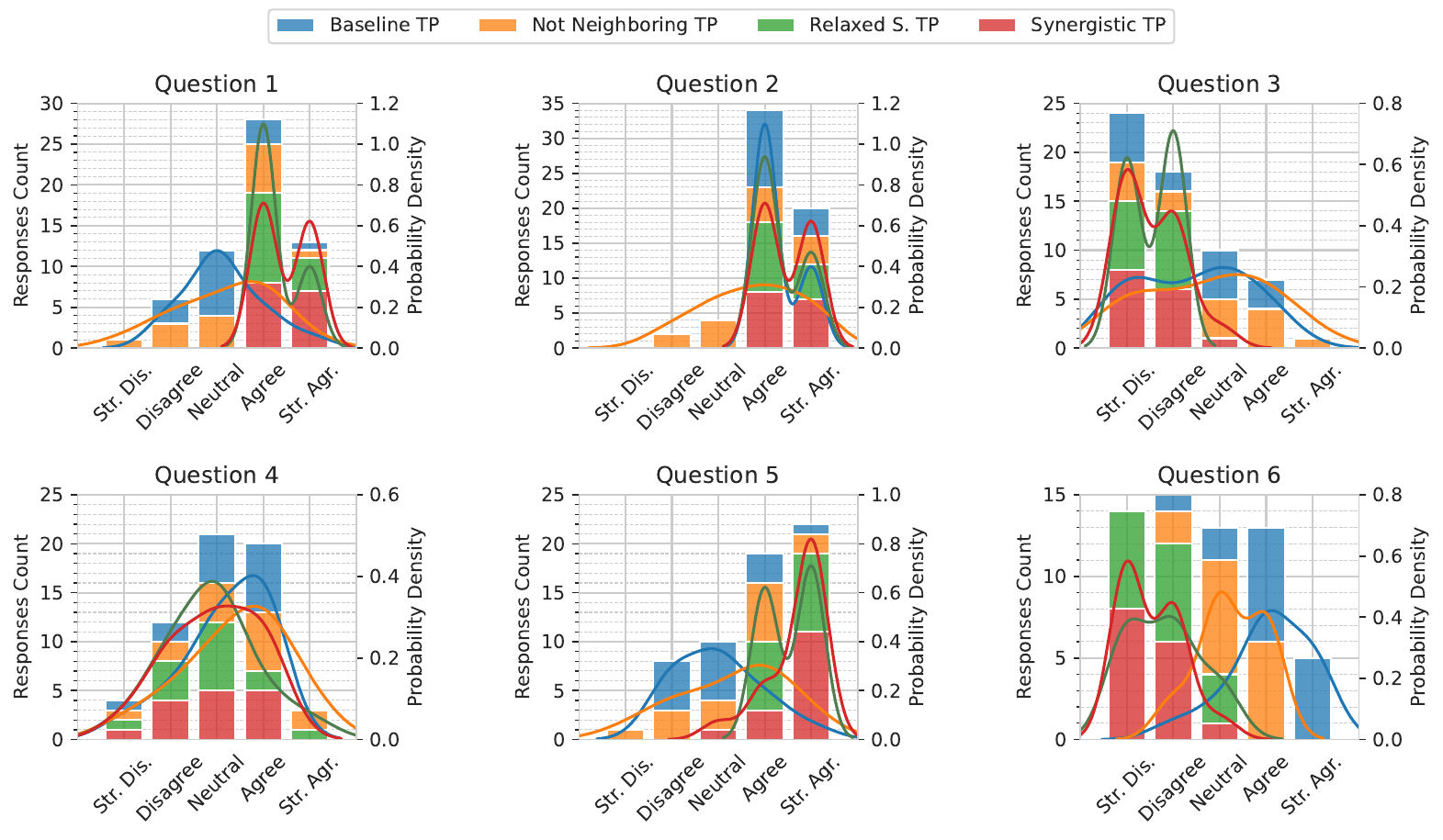}
 \caption{Questionnaire responses for each question. The x-axis displays user responses ranging from ``strongly disagree'' to ``strongly agree''. Responses counts of each method are stacked, with solid lines representing the estimated continuous probability densities. The full text of the questions is provided in Table~\ref{table: questionnaire_statistical_comparison}.}\label{fig:questionnaire_responses}
\end{figure*}

Figure~\ref{fig:questionnaire_responses} displays the user responses to the questionnaire, presenting the response counts for each method in a stacked fashion.
Moreover, a probability density estimation of the user responses is superimposed for each method.
The users' responses were statistically analyzed to test the null hypothesis among the methods: the Kruskal-Wallis Test~\cite{kruskal_wallis} was performed for each item to assess whether statistically significant differences exist among the response populations corresponding to the four methods.
The test results are summarized in the last two columns of Table~\ref{table: questionnaire_statistical_comparison}.
The four task planning methods are statistically equivalent for Q2 (perceived safety) and Q4 (idle time).
The users' responses concerning safety have a median value of ``agree'' for all four methods, and this suggests that the low-level safety module makes the operator feel safe, regardless of the task planning method.
Regarding idle times, the responses have a median value of ``neutral'', except for the Not Neighboring TP, with a median value of ``agree''.
Statistical differences ($p < 0.05$) between methods were observed for overall satisfaction (Q1), perceived human-robot proximity (Q3), fluency (Q5), and robot interruptions (Q6). Additionally, we examined the presence of statistical differences between the pairs of methods we proposed and the comparative methods. Test results are reported in the first four columns of the table ($p$-values), and in all cases, statistical differences ($p < 0.05$) were identified in favor of the proposed methods: 
Q1 has a median value of ``agree'' for the proposed methods and ``neutral'' for the baselines; Q3 has a median of ``strongly disagree'' with STP, ''disagree'' with Relaxed-STP, and ``neutral'' for both baselines; as for Q5 both proposed methods have a median of ``strongly agree'', while Baseline TP scored ``neutral'' and Not Neighbouring TP scored ``agree''; finally, Q6 has a median of ``strongly disagree'' for STP, ``disagree'' for Relaxed-STP, ``neutral'' for Not Neighbouring TP, and ``agree'' for Baseline TP. 

\begin{table*}[htbp]
 \caption{Statistical results of Kruskal-Wallis Test on user's questionnaire responses.
 Method A is the Baseline TP; Method B is the Not Neighboring TP; Method C is the Relaxed-STP; Method D is the Synergistic TP. Cells in the table are highlighted in green if $p$-values$<0.05$ (statistically significant difference detected), in red if statistically equivalent ($p$-values$>0.05$).}\label{table: questionnaire_statistical_comparison}
 \centering

 \begin{tabularx}{\textwidth}{Xcccccc}
 \toprule
 \multirow{2}{*}{Questions} & Method & Method & Method & Method & Statistical & Kruskal-Wallis Test \\
 & C vs A & C vs B & D vs A & D vs B & difference & ($p$-value) \\
 \toprule
 \textbf{Q1}: I am satisfied with the coexistence with the robot to execute the process. & \cellcolor{green!20}2.3e-04 & \cellcolor{green!20}2.2e-03 & \cellcolor{green!20}1.1e-04 & \cellcolor{green!20}6.5e-04 & \cellcolor{green!20}True & \cellcolor{green!20}1.2e-05 \\
 \textbf{Q2}: I felt safe working with the robot. & \cellcolor{red!20}7.0e-01 & \cellcolor{red!20}9.1e-02 & \cellcolor{red!20}2.6e-01 & \cellcolor{green!20}3.8e-02 & \cellcolor{red!20}False & \cellcolor{red!20}1.1e-01 \\[10pt]
 \textbf{Q3}: I felt I was getting too close to the robot to perform my tasks. & \cellcolor{green!20}3.9e-02 & \cellcolor{green!20}1.1e-02 & \cellcolor{green!20}3.9e-02 & \cellcolor{green!20}1.1e-02 & \cellcolor{green!20}True & \cellcolor{green!20}1.1e-02 \\[10pt]
 \textbf{Q4}: Idle times bother me. & \cellcolor{red!20}2.3e-01 & \cellcolor{red!20}1.3e-01 & \cellcolor{red!20}4.0e-01 & \cellcolor{red!20}2.1e-01 & \cellcolor{red!20}False & \cellcolor{red!20}3.5e-01 \\[10pt]
 \textbf{Q5}: The coexistence with the robot during the execution of the process was fluent. & \cellcolor{green!20}8.6e-05 & \cellcolor{green!20}2.1e-03 & \cellcolor{green!20}5.9e-05 & \cellcolor{green!20}7.1e-04 & \cellcolor{green!20}True & \cellcolor{green!20}5.4e-06 \\[10pt]
 \textbf{Q6}: The robot often stopped because you are in the area near the robot & \cellcolor{green!20}1.2e-05 & \cellcolor{green!20}1.1e-04 & \cellcolor{green!20}4.8e-06 & \cellcolor{green!20}1.5e-05 & \cellcolor{green!20}True & \cellcolor{green!20}1.2e-08 \\[10pt]
 \bottomrule
 \end{tabularx}
\end{table*}

\section{Conclusions}

The paper proposes a novel human-aware task allocation and scheduling model based on learned human-robot synergy.
Synergy learning enables the evaluation of coupling effects in concurrent task execution by the agents. Integrating synergy values into the allocation and scheduling model allows for more efficient and safe solutions as early as the task planning stage. The simulations and the case study experiments confirmed the effectiveness of the proposed models. This emerged from both quantitative metrics and subjective user evaluations: the results demonstrate that the proposed methods lead to a reduction in average execution time and a decreased likelihood of small human-robot distances; statistical analysis of user responses revealed a preference for the proposed methods, noting improvements in proximity management, system fluidity, reduced robot interruptions, and overall satisfaction. {
Future works will focus on the integration of risk assessment results into the proposed model: while synergies capture the quantitative impact of human-robot collaboration, the introduction of a penalty matrix, derived from risk evaluations provided by safety experts, would enable the model to account for task-related risks accounting to safety specifications.
Moreover, to transition from offline task planning to replanning, a fast heuristic method for task plan adaptation based on synergies will be investigated to tackle unexpected human behavior.}

\appendix
\section{Linear representation of proposed methods}
\label{appendix: a}

This section presents a linearization of the proposed MINLP models introduced in Section~\ref{sec: methodology}.
The nonlinearity of the model in Section~\ref{sec: expanded_methodology} is primarily related to the cost function, which can be linearized by introducing an auxiliary variable $t^\mathrm{e-max}$, \ie, the makespan, with the following linear constraints:
\begin{equation}\label{makespan_auxiliary}
 \mathcal{C}_{\mathrm{makespan}}:= \quad t^\mathrm{e-max} \geq t^e_i \qquad \forall i \in \left\{1,\dots, m \right\}
\end{equation}
Thus, the nonlinear cost function in~\eqref{makespan_cost_function} is replaced by:
\begin{equation}
 \mathcal{M}=t^\mathrm{e-max}
\end{equation}

The overlapping definition ~\eqref{equ: overlapping} needs to introduce a couple of additional continuous variables (maximum start-time $t^{\mathrm{s-max}}_{i,k}$ and the minimum stop-time $t^{\mathrm{e-min}}_{i,k}$ of the task pair) end three auxiliary binary variables ($\sigma^{\Romannum{1}}_{i,k}$, $\sigma^{\Romannum{2}}_{i,k}$ and $\sigma^{\Romannum{3}}_{i,k}$) for each task couple $(\tau_i,\tau_k)$:
\begin{flalign}
 \mathcal{C}_{\mathrm{max-start}}:= & \notag  \\
 & t^{\mathrm{s-max}}_{i,k} \geq t^s_i \label{t_start_max_1} \\
 & t^{\mathrm{s-max}}_{i,k} \geq t^s_k \label{t_start_max_2} \\
 & t^{\mathrm{s-max}}_{i,k} \leq t^s_i + M (1-\sigma^{\Romannum{1}}_{i,k}) \label{t_start_max_3} \\
 & t^{\mathrm{s-max}}_{i,k} \leq t^s_k + M (\sigma^{\Romannum{1}}_{i,k}) \label{t_start_max_4}
\end{flalign}
\begin{flalign}
 \mathcal{C}_{\mathrm{min-end}}:= & \notag  \\
 & t^{\mathrm{e-min}}_{i,k} \leq t^e_i \label{t_end_min_1} \\
 & t^{\mathrm{e-min}}_{i,k} \leq t^e_k \label{t_end_min_2} \\
 & t^{\mathrm{e-min}}_{i,k} \geq t^e_i - M (1-\sigma^{\Romannum{2}}_{i,k})\label{t_end_min_3} \\
 & t^{\mathrm{e-min}}_{i,k} \geq t^e_k - M (\sigma^{\Romannum{2}}_{i,k})\label{t_end_min_4} 
\end{flalign}

\begin{flalign}
\mathcal{C}_{\mathrm{overlapping}}:= & \notag
 \\
 & \mathrm{OV}_{i,k} \leq (t^{\mathrm{e-min}}_{i,k}-t^{\mathrm{s-max}}_{i,k}) + M \sigma^{\Romannum{3}}_{i,k} \label{OV_1} \\
 & \mathrm{OV}_{i,k} \leq M (1 - \sigma^{\Romannum{3}}_{i,k}) \label{OV_2} \\
 & \mathrm{OV}_{i,k} \geq - M (1-\sigma^{\Romannum{3}}_{i,k}) \label{OV_3} \\
 & \mathrm{OV}_{i,k} \geq 0 \label{OV_4} \\
  & & \mathllap{\forall i,k \in \left\{1,\dots, m\right\} \;, \; k \neq i}\notag
\end{flalign}
Thus,~\eqref{equ: overlapping}-\eqref{delta_time} are replaced by~\eqref{OV_1}-\eqref{OV_4}.
The bilinear term in~\eqref{equ: c_performance} can be linearized as follows:
\begin{equation}
 \label{adapted_t_end}
 t^e_i = t^s_i + \sum_{j \in \mathcal{A}}{\hat{d}^{{j}}_{i} a^j_i} + \sum_{r \in \mathcal{\mathcal{A}^R}}\sum^{m}_{\substack{k = 1 \\ k \neq i}}{\alpha^r_{i,k}(s^r_{i,k}-1)} \quad \forall i \in \left\{1,\dots, m\right\}
\end{equation}
where $\alpha^r_{i,k}$ is an additional binary variable that selects whether or not to apply the additional term to the task duration:
\begin{flalign}
 \mathcal{C}_{\mathrm{adaptation-term}} & := \notag
 \\
 & \alpha^r_{i,k} \leq \mathrm{OV}_{i,k} + M \biggl(2 - a^{H}_{k} - \sum_{j\in\mathcal{A^R}} a^j_i \biggl) \label{alpha_adaptation_1} \\
 & \alpha^r_{i,k} \geq \mathrm{OV}_{i,k} - M \biggl(2 - a^{H}_{k} - \sum_{j\in\mathcal{A^R}} a^j_i \biggl) \label{alpha_adaptation_2} \\
 & \alpha^r_{i,k} \leq M a^{r}_{i} \label{alpha_adaptation_3} \\
 & \alpha^r_{i,k} \geq -M a^{r}_{i} \label{alpha_adaptation_4} \\
 & & \mathllap{\forall i,k \in \left\{1,\dots, m\right\} \;, \; k \neq i\;,\; \forall r \in \mathcal{A}^R}\notag
\end{flalign}

In this case,~\eqref{alpha_adaptation_1}-\eqref{alpha_adaptation_4} are in charge of applying the overlapping effect only if the $(i,k)$ task couple is executed by a robot and a human, respectively.

The same approach can be followed to linearize the model~\ref{sec: relaxed_methodology}. Thus, by adding~\eqref{makespan_auxiliary} to the model, it is possible to rewrite the nonlinear cost function in~\eqref{relaxed_objective} with:
\begin{equation}
 \mathcal{M} = t^{\mathrm{e-max}} + \Delta S
\end{equation}
Moreover, by adding the constraints in~\eqref{t_start_max_1}-\eqref{OV_4} is possible to replace the nonlinear definition of $\Delta S$ in~\eqref{equ: delta_s_multi_robot} with:
\begin{equation}
 \Delta S = \sum_{r \in \mathcal{\mathcal{A}^R}} \sum^{m}_{i=1} \sum^{m}_{\substack{k = 1 \\ k \neq i}}{\alpha^r_{i,k}(s^r_{i,k}-1)}
\end{equation}



\bibliographystyle{cas-model2-names}
\bibliography{bib, new_bib, reference_stiima, references, task_allocation}

\end{document}